\documentclass[10pt,twocolumn,letterpaper]{article}

\usepackage[pagenumbers]{cvpr} % To force page numbers, e.g. for an arXiv version

\usepackage{multirow}

\makeatletter
\def\blfootnote{\xdef\@thefnmark{}\@footnotetext}
\makeatother

\usepackage[dvipsnames]{xcolor}

\definecolor{cvprblue}{rgb}{0.21,0.49,0.74}
\usepackage[pagebackref,breaklinks,colorlinks,citecolor=cvprblue]{hyperref}

\def\paperID{7703} % *** Enter the Paper ID here
\def\confName{CVPR}
\def\confYear{2024}

\title{Image Neural Field Diffusion Models}

\author{Yinbo Chen$^{1}$ \hspace{1mm} Oliver Wang$^{2}$ \hspace{1mm} Richard Zhang$^{3}$ \hspace{1mm} Eli Shechtman$^{3}$ \hspace{1mm}
Xiaolong Wang$^{1\dagger}$ \hspace{1mm} Michael Gharbi$^{3\dagger}$\\ \vspace{-3mm} \\
$^{1}$UC San Diego \hspace{4mm} $^{2}$Google Research \hspace{4mm} $^{3}$Adobe Research \hspace{2mm} \\
\centerline{\href{https://yinboc.github.io/infd/}{https://yinboc.github.io/infd/}}
}

\begin{document}

\twocolumn[{%
 \renewcommand\twocolumn[1][]{#1}%
 \maketitle
 \vspace{-9mm}
 \centering
    \includegraphics[trim={0 25 0 0},width=\linewidth]{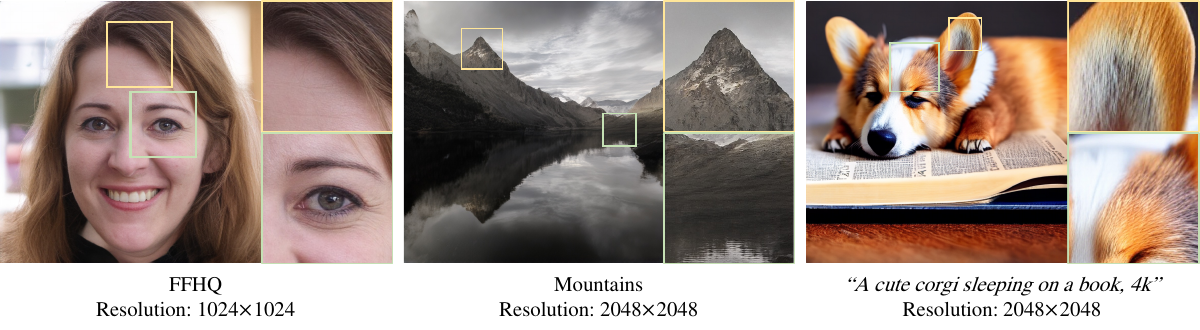}
    \vspace{0.5mm}
    \captionof{figure}{\textbf{Generated samples from our image neural field diffusion models.} We show photorealistic high-resolution image generation by rendering generated image neural fields at 2K resolution for single domain models (left and middle), as well as general text-to-image models (right), with an efficient diffusion process on latent representation at only 64 × 64 resolution.}
    \label{fig:teaser_new}
   \vspace{1mm}
}]

\maketitle

%%%%%%%%% ABSTRACT
\begin{abstract}
    \vspace{-3mm}
    Diffusion models have shown an impressive ability to model complex data distributions, with several key advantages over GANs, such as stable training, better coverage of the training distribution's modes, and the ability to solve inverse problems without extra training.
    However, most diffusion models learn the distribution of fixed-resolution images.
    We propose to learn the distribution of continuous images by training diffusion models on image neural fields, which can be rendered at any resolution, and show its advantages over fixed-resolution models.
    To achieve this, a key challenge is to obtain a latent space that represents photorealistic image neural fields.
    We propose a simple and effective method, inspired by several recent techniques but with key changes to make the image neural fields photorealistic.
    Our method can be used to convert existing latent diffusion autoencoders into image neural field autoencoders.
    We show that image neural field diffusion models can be trained using mixed-resolution image datasets, outperform fixed-resolution diffusion models followed by super-resolution models, and can solve inverse problems with conditions applied at different scales efficiently.
\end{abstract}

%%%%%%%%% BODY TEXT
\vspace{-1em}
\section{Introduction}

\blfootnote{$^{\dagger}$Equal advising.} Diffusion models~\cite{sohl2015deep,ho2020denoising,nichol2021improved} have recently become attractive alternatives to GANs.
These likelihood-based models exhibit fewer artifacts, stable training, can model complex data distributions, do not suffer from mode collapse, and can solve inverse problems using the score function without extra training.
Since diffusion typically requires many iterations at a fixed dimension, directly modeling the diffusion process in the pixel space~\cite{ho2022cascaded,ramesh2022hierarchical,saharia2022photorealistic} can be inefficient for high-resolution image synthesis.
Latent diffusion models (LDMs)~\cite{vahdat2021score,rombach2022high} were proposed as a more efficient alternative. The key idea is to learn an autoencoder to map images to a latent representation from which the image can be decoded back, and train a diffusion model on the lower-dimensional latent representation.
Despite their success, LDMs' latent space still represents images at fixed resolution (for example, 256 in LDM~\cite{rombach2022high} and 512 in Stable Diffusion). 
To generate higher-resolution images (e.g. 2K), LDMs usually first generate a low-resolution image and upsample it using a separate super-resolution model.

In this work, we propose Image Neural Field Diffusion models (INFD). Our method is based on the latent diffusion framework, where we first learn a latent representation that represents an image neural field (which can be rendered at any resolution), then learn a diffusion model on this latent representation.
A key challenge of our approach is to learn a latent space of photorealistic image neural fields where the diffusion model is applied.
We propose a simple and effective method that can convert an existing autoencoder of latent diffusion models to a neural field autoencoder.
We find that directly implementing an autoencoder with LIIF~\cite{chen2021learning} leads to blurred image details, and propose a Convolutional Local Image Function (CLIF), which can render the latent representation to photorealistic high-resolution images and the image content is consistent at different resolutions.
Our neural field autoencoder is trained with L1 loss, perceptual loss~\cite{zhang2018unreasonable}, and GAN loss following LDM~\cite{rombach2022high}, and is supervised from multi-scale patches similar to AnyResGAN~\cite{chai2022any}.

We show that image neural field diffusion models have several key advantages over fixed-resolution diffusion models:
(i) They can be built from mixed-resolution datasets without resizing images. The neural field decoder can render latent representation at any resolution and from patches, which can take supervision from ground-truth images at arbitrary high resolution without decoding the whole image.
(ii) The same latent representation can be supervised with GAN loss from fixed-resolution patches at different scales, with the content consistency across scales, the multi-scale supervision helps high-resolution generation even if all ground-truth images are at a fixed high resolution.
(iii) It does not require an extra Super-Resolution (SR) model for high-resolution generation. Besides the advantage of simplicity, since diffusion-generated low-resolution images do not have high-resolution ground truth, separate SR models are typically trained on real images, while the domain gap between real and generated images could significantly harm the performance of the SR model.
(iv) Image neural field diffusion models learn a resolution-agnostic image prior. Therefore, it can be used to solve inverse problems with a set of conditions defined at different scales efficiently.

In summary, our main contributions are:
\begin{itemize}[nolistsep]
    \item An image neural field autoencoder that can learn representations from mixed-resolution datasets and renders scale-consistent and photorealistic images.
    \item A method to build diffusion models on mixed-resolution datasets and synthesize high-resolution images without extra SR models. Image synthesis is up to 2K resolution with an efficient latent diffusion process at only 64 × 64 resolution (see samples in Figure~\ref{fig:teaser_new}).
    \item A framework to solve inverse problems with conditions applied at different scales of the same image.
\end{itemize}

\begin{figure*}[t]
% \begin{figure*}[t]
    \centering
    \includegraphics[width=0.9\linewidth]{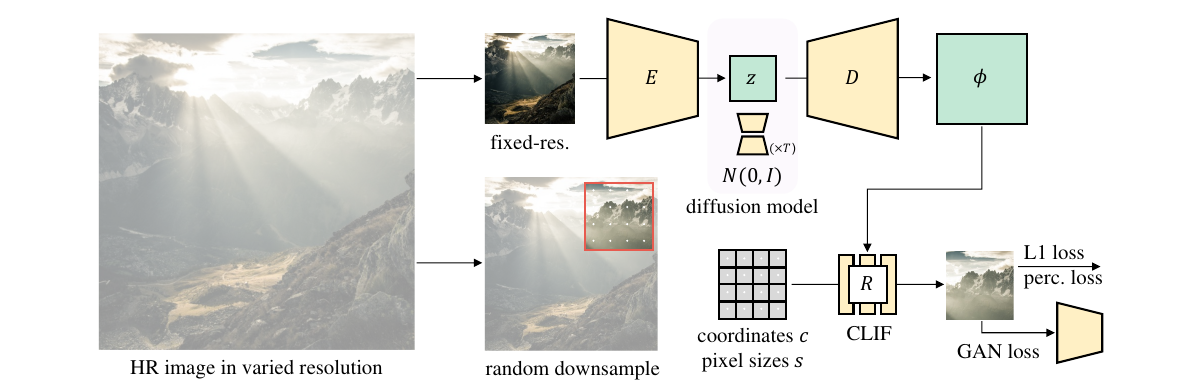}
    \vspace{-3mm}
    \caption{{\bf Method overview.} Given a training image at an arbitrary resolution, we first downsample it to a fixed resolution and pass it into the encoder $E$ to get a latent representation $z$. 
    A decoder $D$ then takes $z$ as input and produces a feature map $\phi$ that drives a neural field renderer $R$, which can render images by querying with the appropriate grid of pixel coordinates $c$ and pixel sizes $s$.
    The autoencoder is trained on crops from a randomly downsampled image ground truth, generating image crops at the corresponding coordinates.
    At test time, a diffusion model generates a latent representation $z$, which is then decoded and used to render a high-resolution image.
    }
    \label{fig:method}
    \vspace{-1em}
% \end{figure*}
\end{figure*}

\section{Related Work}

\paragraph{Diffusion Models.} Diffusion models are first proposed by Sohl-Dickstein et. al.~\cite{sohl2015deep}.
They were recently connected to score-based generative models~\cite{song2019generative,song2020improved,song2020score} and have been greatly improved~\cite{ho2020denoising,nichol2021improved} for architectures and other training details, achieving state-of-the-art results on both unconditional and conditional image synthesis~\cite{saharia2022image,ho2022cascaded,dhariwal2021diffusion,saharia2022palette,nichol2021glide}. 
Compared to prior GAN-based methods~\cite{goodfellow2020generative,radford2015unsupervised,zhu2017unpaired,karras2020analyzing,karras2021alias}, diffusion models have shown nice properties such as stable training, not suffering from mode collapse, and can perform image-to-image translation~\cite{meng2021sdedit,lugmayr2022repaint} or be used to solve inverse problems~\cite{song2021solving}, even with an unconditional model. 
One of the main drawbacks of current diffusion models is slow inference speed, as it relies on iterative reverse diffusion steps. 
While this can be remedied with faster sampling methods~\cite{kong2021fast,san2021noise,lu2022dpm}, performing the diffusion process in the pixel space of high-resolution images remains computationally expensive.

Our work is most closely related to latent diffusion models, which learn to map images to latent representation and train diffusion model on the latent space~\cite{sinha2021d2c,vahdat2021score,rombach2022high}. 
A key design in this direction is to choose the latent representation where the diffusion model is learned. 
Different from prior works that learn an autoencoder, where the latent representation corresponds to a fixed resolution image, we design a decoder and a renderer to learn a latent space that represents image neural fields. 
Since our latent space represents image neural fields, our autoencoder can learn a representation from high-resolution images in varied sizes with multi-scale patches, and can synthesize images at high resolution without relying on extra super-resolution models. 
Our method follows LDM~\cite{rombach2022high}, which trains in two stages.

\vspace{-1em}
\paragraph{Neural Fields for Image Synthesis.} Neural field is also known as Implicit Neural Representations (INR), which represents signals as coordinate-based neural networks. It serves as a compact and powerful differentiable representation and achieves state-of-the-art results mainly for representing 3D shapes~\cite{chen2019learning,park2019deepsdf,mescheder2019occupancy} and scenes~\cite{sitzmann2019scene,saito2019pifu,jiang2020local,peng2020convolutional,chabra2020deep,mildenhall2020nerf,barron2021mip}. Applications of neural fields for images are explored in early works~\cite{stanley2007compositional,mordvintsev2018differentiable} and proposed for more applications such as image super-resolution~\cite{chen2021learning} and image synthesis~\cite{skorokhodov2021adversarial,anokhin2021image}. Several recent works~\cite{karras2021alias,chai2022any} relax the pixel-independent assumption and perform convolutions on the coordinate map to render the output for image synthesis.

The idea of using neural fields for training with any-resolution images is explored in LIIF~\cite{chen2021learning} for autoencoding with an L1 loss, AnyResGAN~\cite{chai2022any} and ScaleParty~\cite{ntavelis2022arbitrary} for GANs with adversarial loss. Our work aims at building the resolution-agnostic learning framework for diffusion models, which is a different model family of generative models.

\vspace{-1em}
\paragraph{Image Super-Resolution.} Image Super-Resolution~\cite{chang2004super,timofte2013anchored,dong2014learning,lim2017enhanced,lai2017deep,zhang2018residual,wang2021real} (SR) aims at upsampling a low-resolution image to higher resolution. Many recent works~\cite{hu2019meta,chen2021learning,yang2021implicit,lee2022local} explore Arbitrary-Scale SR (ASSR) with a single network. While they are related to our method, the differences include: (i) Instead of learning an autoencoder and an extra SR model, our implementation can be viewed as making a bottleneck in a single ASSR model and training diffusion models on the bottleneck. (ii) Our decoder is decoding from latent space to RGB space while super-resolution is upsampling from RGB space to RGB space. (iii) Our input can potentially have any higher resolution information (e.g. crops from high resolution), we choose it as a fixed-low-resolution image for efficiency. IDM~\cite{gao2023implicit} uses a diffusion model for ASSR where the output is at a medium resolution.

\section{Preliminaries}

Our algorithm builds on diffusion models and neural fields. We introduce the core concepts and notation below.

\vspace{-1em}
\paragraph{Diffusion Models.}
Given a sample $x_0$ from a data distribution $q(x_0)$, forward diffusion progressively destroys the information in $x_0$ in $T$ steps, $x_{t-1} \mapsto x_{t}$, each adding some small Gaussian noise.
The process can be concisely rewritten as $x_t = \sqrt{\bar{\alpha}_t} x_0 + \sqrt{1-\bar{\alpha}_t} \epsilon$, where $\epsilon\sim \mathcal{N}(0, I)$, and $\bar{\alpha}_{1\dots T}$ gradually decreases from 1 to 0.
The final distribution is approximately normal, $q(x_T)\sim \mathcal{N}(0, I)$.
A diffusion model learns to reverse this diffusion process.
Once trained, new samples can be generated by first sampling $x_T\sim q(x_T)$ and reversing each step of the diffusion process using a learned transition probability $p_\theta(x_{t-1}| x_t)$, parameterized by a neural network.
Many prior works on image generation are based on maximizing a reweighted variational lower-bound of $p_\theta(x_0)$, which is shown by~\cite{ho2020denoising} to lead to the following training objective:
\begin{equation}
    \mathcal{L} = \mathbb{E}_{x_0,t,\epsilon}\big[||\epsilon_\theta(\sqrt{\bar{\alpha}_t} x_0 + \sqrt{1-\bar{\alpha}_t} \epsilon, t) - \epsilon||^2\big],
\end{equation}
where $t$ is sampled in $\{1,\dots,T\}$, and $\epsilon_\theta$ is the network trained to reverse the diffusion process.

\vspace{-1em}
\paragraph{Neural Fields.}
Neural Fields represent a signal using a coordinate-based neural network.
For example, a neural field can represent an image as a function $c = f(x; \phi)$, where $x\in[-1,1]^2$ are the spatial coordinates in the image domain, $c\in\mathbb{R}^3$ is the RGB color at the corresponding continuous coordinate, and $\phi$ denotes the parameters of the neural field $f$.
Since $x$ can take continuous values, the RGB value can be decoded at arbitrary coordinates.
Accordingly, an image neural field can be rendered at arbitrary resolution by sampling the corresponding pixel coordinates.

\section{Method}

Similar to LDM~\cite{rombach2022high}, our approach has two stages.
First, we train an autoencoder that converts images to latent representations of 2D neural fields (\S~\ref{sec:autoencoder}), which can be rendered to images at any given resolution (\S~\ref{sec:renderer}).
Second, we train a diffusion model to generate samples from this latent space (\S~\ref{sec:diffusion}).
Figure~\ref{fig:method} illustrates our method overview.

\subsection{Image Neural Field Autoencoder}\label{sec:autoencoder}

In the first stage, we seek to convert every image in our training set into a photorealistic image neural field.
We do this by training an autoencoder, made of an encoder $E$, a decoder $D$, and a neural field renderer $R$. 
The encoder maps an RGB input image $I$ to a latent code $z=E(I)$, which is decoded by the decoder to a feature tensor $\phi=D(z)$ used by the neural field renderer to produce the final image.

\vspace{-1em}
\paragraph{Patch-wise decoding.}
For training efficiency, we want to avoid decoding the whole image, because the ground truth can be at a very high resolution.
We take advantage of the coordinate-based decoding property of neural fields, to train with constant-size crops from mixed-resolution data,
which is amenable to batching.
Specifically, we crop a random patch $p_{GT}$ at a fixed $P\times P$ resolution (the red box in Figure~\ref{fig:method}) from a \textit{randomly downsampled} ground-truth.
Since $p_{GT}$ is a fixed-size patch, downsampling the global ground truth lets the patch $p_{GT}$ cover regions at varying scales of the image.
This provides supervision at multiple scales to the latent representation, from local details to global structure.
We discuss this further in \S~\ref{sec:scale_ablation}.
Let $c$ denote the coordinates of pixel centers in the patch within the image, and $s$ denote their pixel sizes relative to the whole ground-truth image.
Our renderer $R$ takes as input the features $\phi=D(z)$ decoded from the latent representation, and the coordinates and pixel sizes $c,s$ to synthesize an output patch $p=R(c, s;\phi)$. 

\vspace{-1em}
\paragraph{Training objective.}
We compare the synthesized patch to the ground truth using a sum of an $L_1$ loss, a perceptual loss $\mathcal{L}_\text{perc}$~\cite{zhang2018unreasonable}, and an adversarial loss $\mathcal{L}_{\text{GAN}}$~\cite{goodfellow2020generative}.
The discriminator is simultaneously trained to distinguish between the distributions of $p$ and $p_{GT}$.
Thus, we minimize the following objective to train $E$, $D$, and $R$:
\begin{equation}
    \mathcal{L}_{AE} = ||p - p_{GT}||_1 + \mathcal{L}_\text{perc}(p, p_{GT}) + \mathcal{L}_{\text{GAN}}(p).
\end{equation}

\begin{figure}
    \centering
    \includegraphics[width=0.8\linewidth]{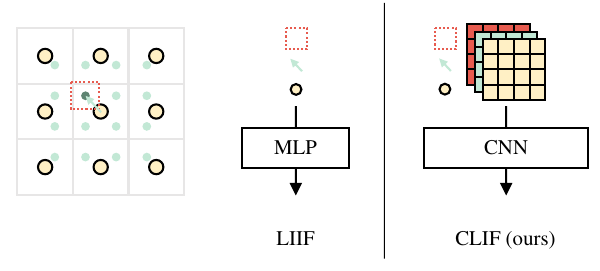}
    \vspace{-4mm}
    \caption{{\bf Convolutional Local Image Function (CLIF).} Given a feature map $\phi$ (yellow dots), for each query point $x$ (green dot), we fetch the nearest feature vector, along with the relative coordinates and the pixel size. The grid of query information is then passed into a convolutional network (right) that renders an RGB grid. Different than the pointwise function LIIF, CLIF has a higher generation capacity and is learned to be still scale-consistent.}
    \label{fig:conv-liif}
    \vspace{-1em}
\end{figure}

Our implementation follows the autoencoder architecture of LDM~\cite{rombach2022high}, with the same encoder and decoder (removed the last layer) architectures to facilitate comparisons.
Since training images have arbitrary resolutions, we resample the encoder's input to a fixed resolution $256\times 256$.
Note that despite this downsampling, we still train against mixed-resolution references.
The encoder and decoder are with a spatial downsampling/upsampling rate of 4 correspondingly, therefore the latent representation is in $64\times 64$.
A vector-quantization (VQ) layer is prepended to the first layer of the decoder to regularize the latent space. $\phi$ is $256\times 256$ with $128$ channels. We set patch size $P=256$.

\subsection{Neural Field Renderer}\label{sec:renderer}

Our renderer $R$, shown in Figure~\ref{fig:conv-liif}, is a neural field coordinate-based decoder, which we dubbed Convolutional Local Image Function (CLIF).
To decode an image patch with CLIF, each query point $c$ (green dot) fetches the spatially nearest feature vector from the feature map $\phi$ (yellow dots).
We concatenate the nearest feature vector with the query coordinates $c$ and pixel size $s$, then process the grid of query information using a convolutional network to output an RGB image.
Intuitively, the concatenated feature is the field information at a point.
By changing the query coordinates and pixel sizes, we can decode images at any resolution.
LIIF~\cite{chen2021learning} decodes with similar information, but it uses a pointwise function.
We found this limits LIIF's ability to produce realistic high-frequency details (see Appendix~\ref{sec:cmp_liif_ae}). 
CLIF remedies this issue by exploiting more local feature context.
Our CLIF renderer is learned to be scale-consistent, i.e., details are consistent when decoding at different resolutions (see \S~\ref{sec:scale_consistent}).

\begin{figure*}
    \centering
    \includegraphics[width=\linewidth]{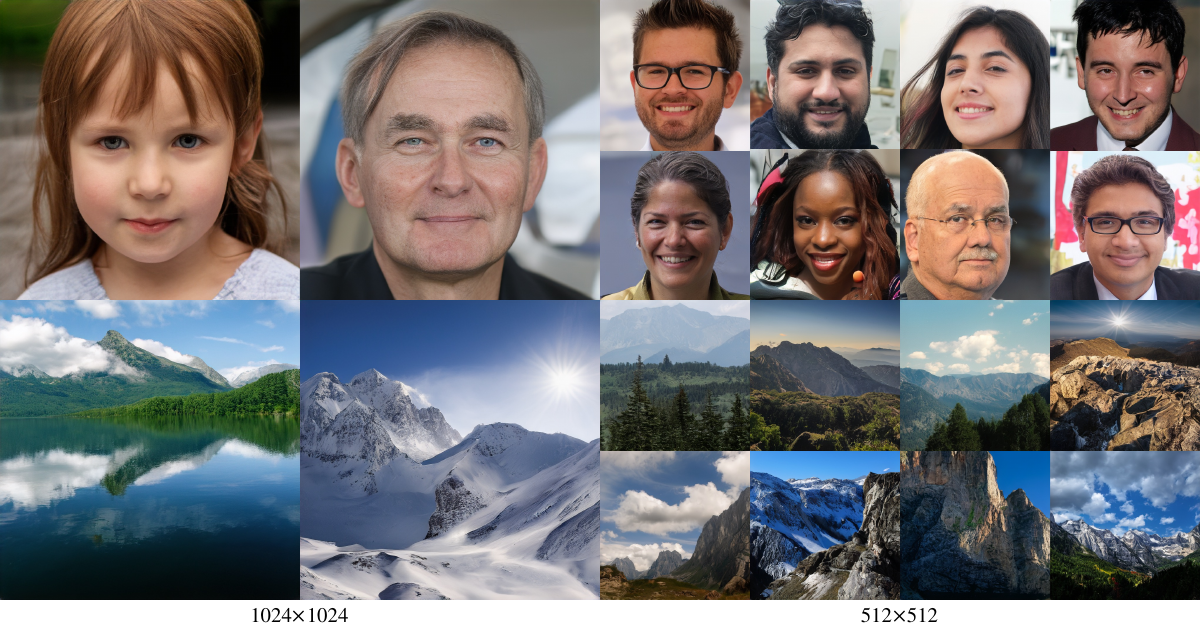}
    \vspace{-2em}
    \caption{Generated samples from our method on FFHQ and Mountains dataset.}
    \vspace{-1em}
    \label{fig:samples_ffhq}
\end{figure*}

\subsection{Latent diffusion}\label{sec:diffusion}
Once the autoencoder is trained, we map every image $I$ in the training dataset to its latent representation $z$, and train a diffusion model by optimizing the DDPM~\cite{ho2020denoising} objective
\begin{equation}
    \mathcal{L}_{DM} = \mathbb{E}_{z\sim E(I),t,\epsilon}\big[||\epsilon_\theta(\sqrt{\bar{\alpha}_t} z + \sqrt{1-\bar{\alpha}_t} \epsilon, t) - \epsilon||^2\big],
\end{equation}
using the distribution over $z$ induced by the encoder.
After training, the encoder can be discarded. 
The diffusion model generates a latent representation $z$, which is then decoded to $\phi=D(z)$ and rendered at a resolution specified by the pixel coordinates $R(c, s;\phi)$ as described next.

\subsection{Patchwise Image Generation}

Despite being trained with small patches for efficiency, our method can generate high-resolution images.
For this, we first generate a global feature map $\phi$ from a sampled $z$, then generate sub-tiles of a large image by querying the renderer at the corresponding coordinates, as described in \S~\ref{sec:renderer}.
To avoid discontinuities at tile boundaries, we expand the query region for each tile by a fixed padding size larger than CLIF's receptive field (8 in our experiments).
We then crop the output tiles by the same amount and assemble the tiles into a seamless composite.
Our renderer is fully convolutional, it can also generate the image at once, as long as memory is sufficient to hold intermediate buffers.

\section{Experiments}

We evaluate our method on several datasets and compare it to LDM~\cite{rombach2022high} with super-resolution models (\S~\ref{sec:ldm_comparison}). Sections \S~\ref{sec:scale_ablation} and \S~\ref{sec:hr_ablation} presents model ablations. We show results in solving multi-scale inverse problems in \S~\ref{sec:inverse}, and text-to-image generation in \S~\ref{sec:text2im}. We further discuss the comparison with prior any-resolution GANs in \S~\ref{sec:cmp_anyresgan}.

\vspace{-1em}
\paragraph{Data.}
The FFHQ~\cite{karras2019style} used in LDM~\cite{rombach2022high} contains 70K high-resolution images ($1024\times 1024$) and several baselines use it for comparison.
When comparing to LDM, we follow their training and validation split: 60K images for training and 10K for validation.
Since our method is flexible and not limited to a fixed-resolution dataset, we also follow a controlled setting in prior work~\cite{chai2022any}, which constructs a varied-resolution dataset from FFHQ by constructing and merging three sets: (i) all images at 256 low-resolution; (ii) a subset of 5K samples in varied-resolution from 512 to 1024; (iii) a subset of 1K images at 1024 resolution.
We denote this setting as FFHQ 6K-Mix.
Besides FFHQ, we evaluate our method on the Mountains dataset~\cite{chai2022any}, which contains a low-resolution subset, with about 500K samples around 1024 resolution,
and a high-resolution subset, with about 9K images at resolutions beyond 2048.
Figure~\ref{fig:samples_ffhq} shows examples of our generated results on these datasets.

\vspace{-1em}
\paragraph{pFID metric.}
Standard FID evaluation first resizes images to 299, which evaluates the global structure of images regardless of their original resolution, but is insufficient to measure the quality of details from high-resolution generators.
Patch-FID~\cite{chai2022any} (pFID) addresses this limitation.
It computes the FID between fixed-resolution patches cropped from arbitrary resolution ground-truth and generated images, thus evaluating local details in synthesized images.

The original pFID reports the metric on random crops from images at varying resolutions.
Since our practical focus is on high-resolution synthesis, we evaluate pFID between patches cropped from fixed-high-resolution images. 
To further disentangle the evaluation for image details at different scales, we separately report pFID for patches at different resolutions.
$P$/1K denotes the FID between random crops at resolution $P$ from ground truth and synthesized images at resolution 1024.
For example, pFID-256/1K evaluates the local details, pFID-1K/1K evaluates the global structure and is the same as standard FID applied to 1024-resolution images.
We generate 50K samples to compute FID in most experiments.
In some experiments, we use 5K samples if it suffices to observe the performance gap (specified after FID@ in Tables).

\begin{figure*}
    \centering
    \includegraphics[width=\linewidth]{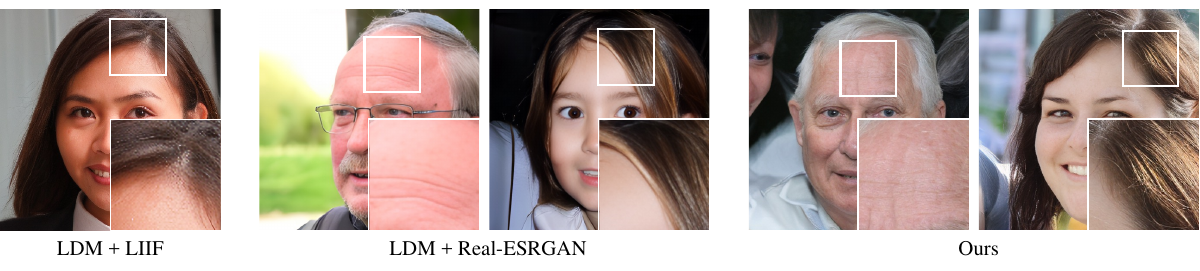}
    \vspace{-7mm}
    \caption{Qualitative comparison with LDM followed by super-resolution on FFHQ. LIIF shows noise in details, while Real-ESRGAN tends to be smooth and results lack rich details. Our approach generates images with more realistic details.}
    \label{fig:cmp_ldm_sr}
\end{figure*}

\begin{table*}[]
    \small
    \centering
    \begin{tabular}{cccccccc}
        \toprule
        \multirow{2}{*}{\textbf{Model}} & \multirow{2}{*}{\textbf{\#Params}} & \multirow{2}{*}{\textbf{Coordinate-based Decoder}} & \multicolumn{4}{c}{\textbf{pFID@50K}} \\
        \cmidrule(lr){4-7}
         & & & 256/256 & 256/1K & 512/1K & 1K/1K \\
        \midrule
        LDM~\cite{rombach2022high} & 33.0M & & 4.98 & - & - & - \\
        LDM + LIIF~\cite{chen2021learning} & 33.0M + 22.3M & \checkmark  & - & 56.65 & 17.83 & 8.97 \\
        LDM + ITSRN-RDN~\cite{yang2021implicit} & 33.0M + 22.6M & \checkmark  & - & 51.73 & 17.94 & 8.02 \\
        LDM + SwinIR-LTE~\cite{lee2022local} & 33.0M + 12.1M & \checkmark  & - & 52.62 & 17.72 & 9.09 \\
        LDM + Real-ESRGAN~\cite{wang2021real} & 33.0M + 16.7M &  & - & 18.38 & 18.46 & 16.04 \\
        \midrule
        LDM (our reimplementation) & 33.0M &  & 6.02 & - & - & - \\
        \midrule
        INFD (ours) & 35.7M & \checkmark & 5.34 & \textbf{8.07} & \textbf{6.64} & \textbf{5.57} \\
        \bottomrule
    \end{tabular}
    \caption{Comparison to Latent Diffusion Model (LDM) with super-resolution models for high-resolution image synthesis on FFHQ dataset. \#Params counts for the decoder, and the renderer or super-resolution model if it exists, which are used in image generation.
    }
    \vspace{-1em}
    \label{tab:ldm}
\end{table*}

\subsection{Comparison to LDM}\label{sec:ldm_comparison}

LDM~\cite{rombach2022high} trains an autoencoder for images at $256\times 256$ resolution and learns a diffusion model on its latent space.
For a fair comparison, we used the same encoder and decoder architectures as LDM.
Our method only adds a lightweight CLIF renderer, which contains 2 convolution layers and 2 ResNet~\cite{he2016deep} blocks.

Even though LDM's diffusion network is fully convolutional and can generate high-resolution images by diffusing from a larger noise map, it is known that this approach generates repetitive patterns (e.g. distorted faces with duplicate features, see Appendix~\ref{sec:ldm_conv_samples}).
As a result, we follow the standard approach to generate high-resolution images with LDM by running an independent super-resolution model on its output.
We combine LDM with several recent state-of-the-art arbitrary-scale super-resolution methods: LIIF~\cite{chen2021learning}, ITSRN~\cite{yang2021implicit}, LTE~\cite{lee2022local}, which allow for inference at continuous upsampling scales, and Real-ESRGAN~\cite{wang2021real}, a super-resolution model for a fixed upsampling scale that has state-of-the-art perceptual quality.
We report qualitative results in Figure~\ref{fig:cmp_ldm_sr}, and the pFID in Table~\ref{tab:ldm} on FFHQ. As Real-ESRGAN shows the most competitive results, we also compare to it on Mountains dataset in Table~\ref{tab:ldm_mountains}. 

We find that for standard FID at 256 resolution (i.e. pFID 256/256), the image neural field diffusion model is competitive with the original LDM, which can only generate images at 256.
Our retraining of LDM reaches a slightly worse FID than the officially reported scores, which we attribute to implementation differences and training variance.
At all higher resolutions, our method outperforms LDM followed by super-resolution, which is consistent with our qualitative observation and suggests we achieve better image quality for high-resolution detail at different scales.
We hypothesize that the main issues for extra super-resolution models are that: (i) LDM with an extra super-resolution model can be viewed as first decoding the latent representation to RGB space, then upsample to another RGB space, where the first RGB space becomes a bottleneck that contains much less information than feature space; (ii) the artifacts generated by LDM, even inconspicuous, can cause a domain shift to the input of the super-resolution models, which could significantly degrade their performance. The super-resolution models can not be directly trained to upsample the generated low-resolution images since no paired high-resolution ground truth is available. In our method, the domain shift from real to generated samples happens in the latent space (with VQ or KL regularization). We hypothesize that latent space is more robust than RGB space to the domain shift. We observe that when replacing the latent space with RGB space, the generated images become overly smooth similar to LDM with Real-ESRGAN super-resolution, which is consistent with our hypothesis (see Appendix~\ref{sec:without_latent_space}).

\subsection{Effect of scale-varied training}
\label{sec:scale_ablation}

Randomly downsampling the global image before extracting fixed-resolution training patches would make patches cover all scales. With the scale consistency of CLIF, an image neural field is supervised to be realistic in all scales via the perceptual and GAN losses.
We conduct an ablation on FFHQ to evaluate the impact of this random downsampling strategy.
Specifically, we disable it during training, keeping all ground-truth images at 1024 resolution. The results are shown in Table~\ref{tab:abl_down}.
Without random downsampling, the pFID is worse especially for 512/1K and 1K/1K, suggesting that random downsampling of the ground truth improves quality, even if we only aim at generating images at 1024 high resolution, because it helps supervise every single image to be realistic at all scales.
For Mountains dataset, we observe that there will be obvious artifacts without random downsampling (see Appendix~\ref{sec:scale_varied_mountains}).
This contrasts with the observation in LIIF~\cite{chen2021learning}, which found that random downsampling hurts performance at a fixed highest scale when it is only using an L1 loss.

\begin{table}[]
    \centering
    \small
    \begin{tabular}{cccc}
        \toprule
        \multirow{2}{*}{\textbf{Model}} & \multicolumn{3}{c}{\textbf{pFID@50K}} \\
        \cmidrule(lr){2-4}
         & 256/1K & 512/1K & 1K/1K \\
        \midrule
        LDM + Real-ESRGAN~\cite{wang2021real} & 17.36 & 15.11 & 10.39 \\
        INFD (ours) & \textbf{7.53} & \textbf{6.84} & \textbf{5.13} \\
        \bottomrule
    \end{tabular}
    \caption{Comparison to Latent Diffusion Model (LDM) with super-resolution models on Mountains dataset.}
    \vspace{-1.5em}
    \label{tab:ldm_mountains}
\end{table}

\subsection{Training with limited high-resolution images}\label{sec:hr_ablation}

While previous experiments use high-resolution training images, a key advantage of 
our method is that it can learn from mixed-resolution datasets and still generate high-resolution outputs, even when the number of high-resolution images is limited. 
We quantify this in Table~\ref{tab:abl_mix} using FFHQ 6K-Mix, where most images are at low resolution 256, 5K images are at 512--1024, and 1K images (1.4\% of the dataset) are at 1024. 
Training with the mixed-resolution dataset as-is (also with random downsampling) already yields a model that performs decently at 1024, but with worse details than the model trained with all images at 1024. 
We suspect that the model is optimized for too few steps using images at 1024, since the number of high-resolution images is small and they go through random downsampling. 
We balance the training resolution by sampling images from the 5K + 1K high-resolution subset with a probability of 0.5.
This largely closes the performance gap (6K-Mix, bal.) with the model train on high-resolution images only.

\begin{table}[]
    \centering
    \small
    \begin{tabular}{cccccc}
        \toprule
        \multirow{2}{*}{\textbf{Downsample}} & \multicolumn{3}{c}{\textbf{pFID@50K}} \\
        \cmidrule(lr){2-4}
         & 256/1K & 512/1K & 1K/1K \\
        \midrule
        Fix 1024 & 8.19 & 6.82 & 6.04 \\
        256--1024 & \textbf{8.07} & \textbf{6.64} & \textbf{5.57} \\
        \midrule
        $\Delta$ & 0.12 & 0.18 & 0.47 \\
        \bottomrule
    \end{tabular}
    \vspace{-0.5em}
    \caption{FFHQ dataset. Random downsampling during training improves image generation quality even for fixed high resolution.}
    \vspace{-1em}
    \label{tab:abl_down}
\end{table}

\subsection{Image generation beyond 1024}

We explore going beyond the 1024 resolution and train our model on a collected dataset of faces at varied resolutions between 1024 to 2048, and on Mountains dataset including a higher-resolution subset (a generated sample is shown in Figure~\ref{fig:teaser_new}). We observe that our method can be effectively applied to resolutions up to 2K (see Appendix~\ref{sec:gsfaces_results}).

\begin{table}[]
    \centering
    \small
    \begin{tabular}{cccccc}
        \toprule
        \multirow{2}{*}{\textbf{Data}} & \multicolumn{3}{c}{\textbf{pFID@50K}} \\
        \cmidrule(lr){2-4}
         & 256/1K & 512/1K & 1K/1K \\
        \midrule
        All HR & 8.07 & 6.64 & 5.57 \\
        \midrule
        6K-Mix & 16.27 & 10.8 & 11.3 \\
        6K-Mix, bal. & 12.41 & 7.74 & 6.9 \\
        \bottomrule
    \end{tabular}
    \vspace{-0.5em}
    \caption{FFHQ dataset. Our method can learn from mixed-resolution images with a limited number of full-resolution images.}
    \vspace{-1em}
    \label{tab:abl_mix}
\end{table}

\begin{figure}
    \centering
    \captionsetup{type=figure}
    \includegraphics[trim={0 25 0 0},width=\linewidth]{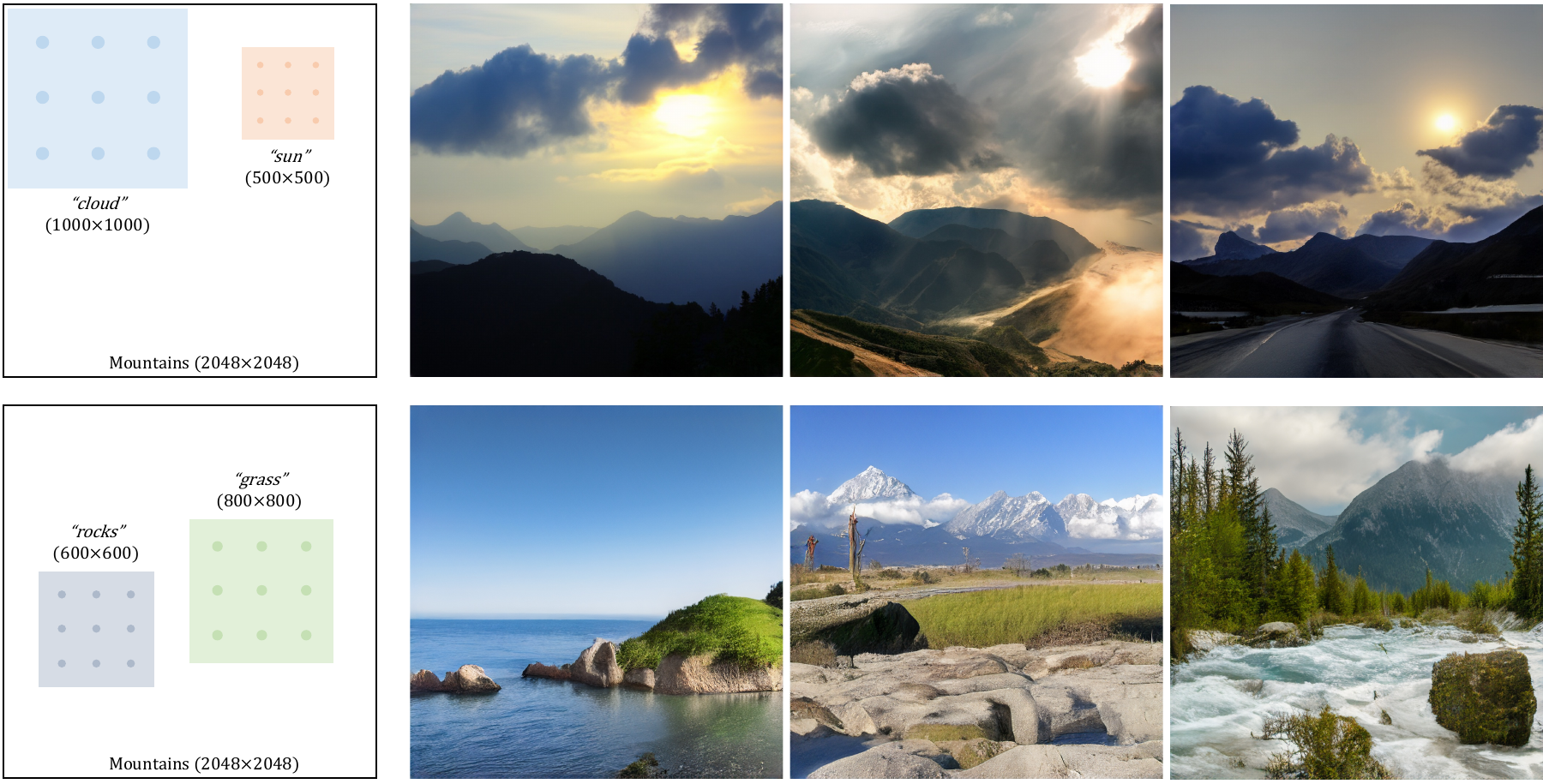}
    \vspace{-4mm}
    \captionof{figure}{\textbf{Solving inverse problems with multi-scale conditions per image.}
    We can solve for an image that satisfies multi-scale conditions, defined as square regions and a text prompt (left).
    For this, we decode the corresponding region and pass it to a pre-trained CLIP~\cite{radford2021learning} model operating at fixed-resolution ($224\times 224$),
    and maximize the CLIP similarity to the corresponding text prompt.
    This enables layout-to-image generation without extra training. We show generated solutions on the right.}
    \label{fig:teaser_inverse}
    \vspace{-1em}
\end{figure}

\begin{figure*}[t]
    \centering
    \includegraphics[width=\linewidth]{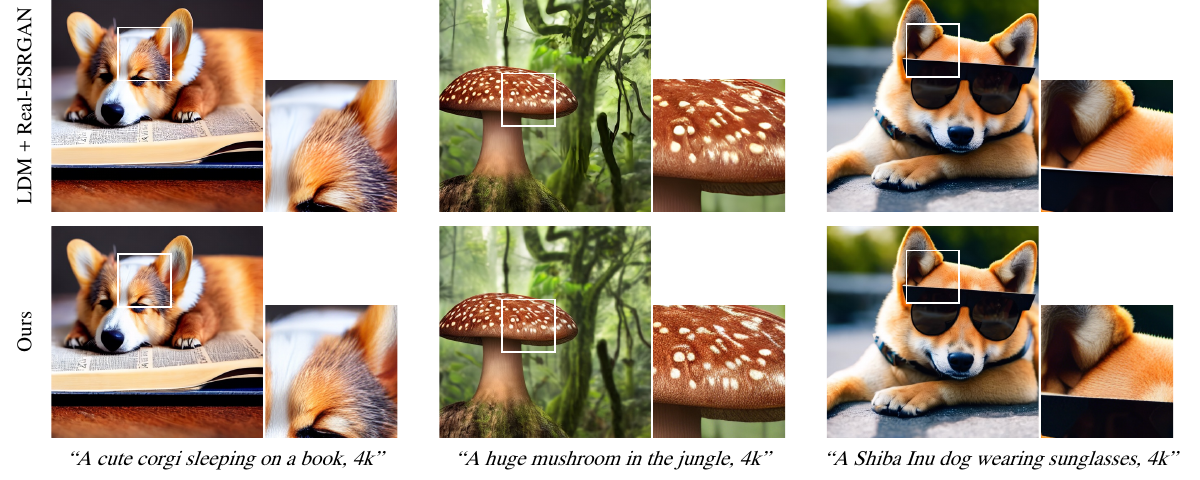}
    \vspace{-6mm}
    \caption{Samples of our method finetuned from Stable Diffusion (LDM) compared to Stable Diffusion with an extra super-resolution model. Our approach yields finer, high-frequency textures.}
    \label{fig:txt2img}
    \vspace{-1em}
\end{figure*}

\subsection{Inverse problems with conditions at any scale}
\label{sec:inverse}

Our method builds a diffusion model on a latent space that represents image neural fields. A key property of image neural fields is that they can be efficiently rendered for any sub-region at any given resolution without decoding the whole image at full resolution. With the image prior learned by diffusion models, it enables efficiently solving inverse problems where conditions can be defined on any scale of the image based on coordinates. We take zero-shot any-scale layout-to-image generation as an example. It uses CLIP~\cite{radford2021learning} similarity as the constraint for image generation with semantic bounding boxes at arbitrary scales.

Specifically, we take a pre-trained CLIP model, which takes $224\times 224$ fixed-resolution images as inputs. Given a layout and our image neural field diffusion model (unconditional), in each diffusion step, let $z$ denote the current denoised latent representation, for each semantic bounding box $i$ in the input layout we render $z$ for the corresponding sub-region to a patch at the resolution $224\times 224$, i.e. patch $p_i = R\circ D(z; c_i, s_i)$, with our decoder $D$ and renderer $R$, where $c_i, s_i$ denote the coordinates and scale of a $224\times 224$ pixel grid of bounding box $i$. The clip similarity loss $l_i = \textrm{CLIP}(p_i, T_i)$ is computed between patch $p_i$ and the given text $T_i$. The gradients $\frac{\partial l_i}{\partial z}$ are back-propagated and then used to modify the diffusion score for each diffusion step. We follow the techniques in DPS~\cite{chung2023diffusion} as the inverse problem solver. Figure~\ref{fig:teaser_inverse} shows the results of using our mountains model (2K resolution). Note that without image neural field diffusion models, a fixed-resolution diffusion model needs to decode the whole region at full resolution (e.g. 1000 × 1000 ``cloud'' region in a 2048 × 2048 canvas) before passing it as 224 × 224 input to CLIP, which would have very intensive computation and memory cost.

\subsection{Qualitative results for text to image generation}\label{sec:text2im}
\label{sec:txt2img}

We explore a preliminary application of our method for text-to-image generation by finetuning a pre-trained Stable Diffusion model~\cite{rombach2022high}.
Because of the high computational cost of training Stable Diffusion, we freeze the encoder and only finetune the decoder from
publicly available pre-trained weights on a high-resolution subset of LAION-5B, which contains samples at resolutions higher than 2K.
Our renderer has the same architecture as in previous experiments and is jointly trained from scratch. 
We show some qualitative samples from our fine-tuned model and compare them to upsampling the Stable Diffusion's 512$\times$512 output using Real-ESRGAN in Figure~\ref{fig:txt2img}. 
We observe that the comparison is similar to the experiments on FFHQ and Mountains: our method generates more details than the Real-ESRGAN applied to Stable Diffusion's outputs (see Appendix~\ref{sec:additional_samples}).

\subsection{Scale-consistency of CLIF}
\label{sec:scale_consistent}

The CLIF neural field renderer does not assume pixel independence and is learned with LPIPS and GAN loss. However, instead of synthesizing different contents for different output scales as a conditional GAN, we observe that CLIF is learned to be scale-consistent even without any explicit consistency objective, as shown in Figure~\ref{fig:scale_consistency}. We observe that the object boundary precisely aligns when we render the same latent representation to different resolutions. This property can help multi-scale supervision on the same latent representation in training and solving inverse problems.

\subsection{Comparison to any-resolution GANs}
\label{sec:cmp_anyresgan}

The idea of using neural fields for training with any-resolution images was explored for GANs in AnyResGAN~\cite{chai2022any} and ScaleParty~\cite{ntavelis2022arbitrary}.
We observe that the comparison between image neural field diffusion models and AnyresGAN matches the comparison between typical fixed-resolution diffusion models and GANs.
We take AnyResGAN~\cite{chai2022any}, which worked for more diverse high-resolution images, as an example for comparison.
GANs are still state-of-the-art on FID, especially for the single-class generation. For example, for typical fixed-resolution synthesis on FFHQ, StyleGANv3~\cite{karras2021alias} (the backbone of AnyResGAN) reports FID 2.79 while LDM~\cite{rombach2022high} (the backbone of our implementation) has FID at 4.98. However, FID is counting for the statistics of deep features and is not a perfect metric yet for image generation~\cite{borji2019pros,borji2022pros}. Image neural field diffusion model inherits the advantages of diffusion models over GANs. We detail below.
% It is also learned to be scale-consistent.

\begin{figure*}
    \centering
    \includegraphics[width=\linewidth]{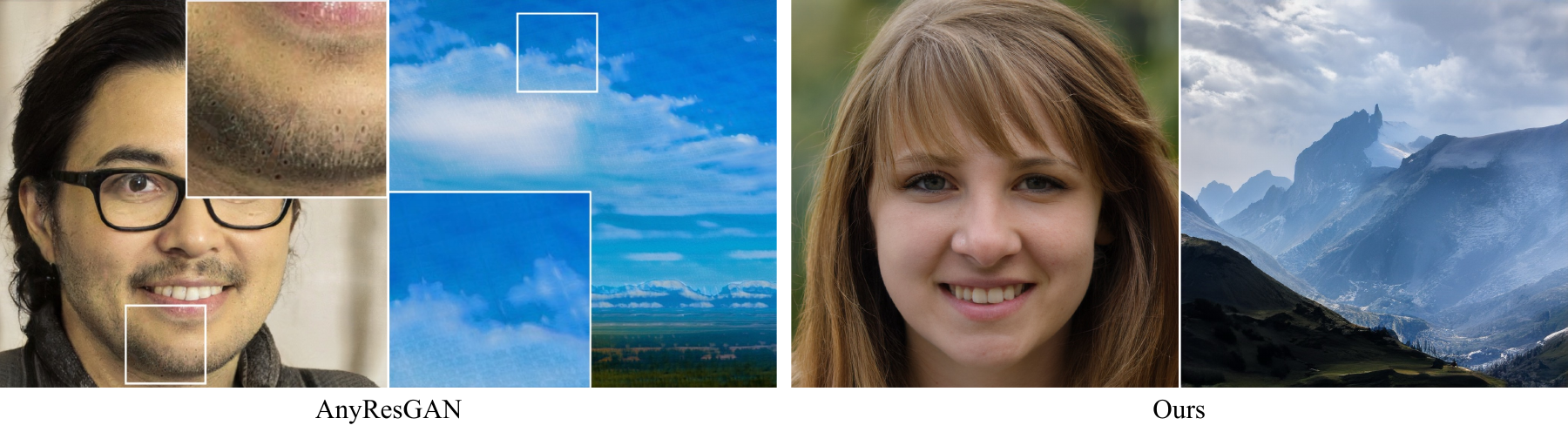}
    \caption{Qualitative comparison with AnyResGAN~\cite{chai2022any}. Image neural field diffusion model is based on diffusion models and avoids the GAN artifacts in FFHQ (black dots) and Mountains (grid-like patterns) (see more examples in Figure~\ref{fig:samples_ffhq}).}
    % \vspace{-1em}
    \label{fig:anyresgan}
\end{figure*}

\paragraph{Sample quality.} While our method achieves competitive FID to AnyResGAN (see Appendix Table~\ref{tab:cmp_anyres}), as a diffusion model, we find it shows better quality in actual samples.
We observe that the common artifacts in AnyResGAN (see Figure~\ref{fig:anyresgan}) are not shown in image neural field diffusion. We also find the generally best samples from our method have better quality than the best samples from AnyResGAN.

\paragraph{Sample diversity.}
We visualize random samples from ground-truth images in Appendix Figure~\ref{fig:sample_diversity_gt}, AnyResGAN in Appendix Figure~\ref{fig:sample_diversity_gan}, and image neural field diffusion model in Appendix Figure~\ref{fig:sample_diversity_dm}. Overall, we observe that AnyResGAN's samples layout are relatively more flat and simple (typically a front view of a mountain, with a horizontal sky-mountain line), while the diffusion-based method has better sample diversity (usually more layers and contents along the depth in the layout).

Besides, unlike AnyResGAN, INFD is scale-consistent as shown in Figure~\ref{fig:scale_consistency}, and can be used for text-to-image synthesis (see more samples in Appendix Figures~\ref{fig:add_txt2img2},\ref{fig:add_txt2img3},\ref{fig:add_txt2img1}), which remains a challenge for any-resolution GANs.

\section{Limitations}

\begin{figure*}
    \centering
    \includegraphics[width=\linewidth]{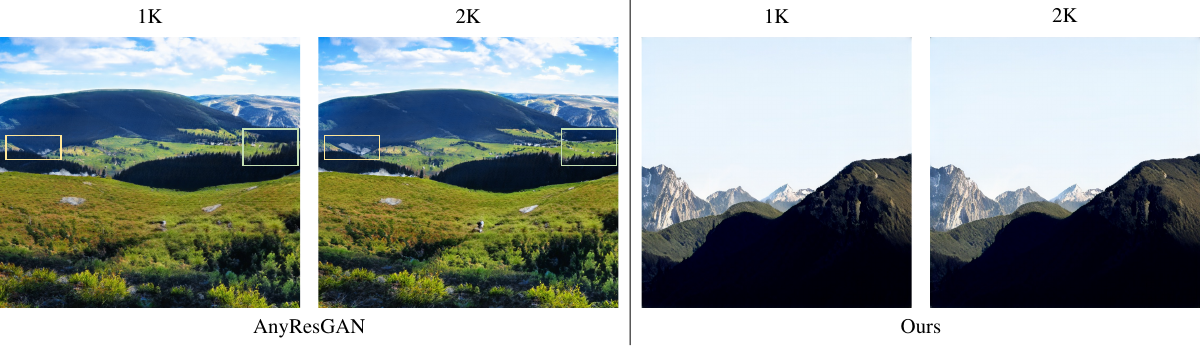}
    \caption{Scale consistency of our CLIF renderer. In each pair, the left is rendering at resolution 1K, the right is rendering at resolution 2K then downsampling to 1K. Yellow/green boxes show two examples of inconsistent areas in AnyResGAN (besides the boxes, the object boundary also does not align well for 1K/2K of AnyResGAN).}
    \label{fig:scale_consistency}
\end{figure*}

Our current method assumes that the training data are scale-consistent, i.e., low-resolution images follow the same distribution as downsampled high-resolution images (see Appendix~\ref{sec:dataset_scale_consistency}). 
This assumption is violated by datasets that contain a low-resolution subset with noisy, compressed images, and a high-resolution subset with clean images (e.g. Birds, Churches datasets~\cite{chai2022any}).

In text-to-image synthesis, our model is fine-tuned from an existing Stable Diffusion checkpoint on a small subset of the LAION dataset containing high-resolution images.
This causes two issues.
First, the training set of the pre-trained model includes noisy images, but our high-resolution fine-tuning dataset only contains clean images; this violates our scale consistency assumption.
As a result, we found that our model requires extra prompts such as ``4k'' to generate detailed high-resolution images.
Second, our fine-tuning LAION subset does not cover all possible object categories, so our model may not perform optimally on some out-of-distribution objects.
Training from scratch on the full LAION dataset might resolve these limitations.

Our current encoder works for images at a fixed resolution. Researching efficient any-resolution encoders is also a promising avenue for future work.
At very high resolutions, our model sometimes generates artifacts in high-frequency regions. We hypothesize this because the current LPIPS and GAN loss are not optimal for model uncertainties. Our current CLIF implementation is also lightweight, which may have a limited representation capacity.

\section{Conclusion}
We proposed image neural field diffusion models, the diffusion models on a resolution-agnostic latent space, and demonstrated its advantages over fixed-resolution models.
We presented a simple yet effective framework as an implementation, which can be easily applied to convert from an existing latent diffusion model.
Our method can build diffusion models from mixed-resolution datasets, achieving high-resolution image synthesis without extra super-resolution models.

The resolution-agnostic image prior learned by the diffusion model also enables solving inverse problems with conditions applied at different scales of the same image.
The image neural field can be rendered by patches as needed to efficiently compute the constraint loss and solve for very high-resolution images.

\subsection*{Acknowledgements} This work was supported, in part, by NSF CAREER Award IIS-2240014, and the CISCO Faculty Award.

{
    \small
    \bibliographystyle{ieeenat_fullname}
    \bibliography{main}
}

%%%%%%%%% BODY TEXT

\appendix
\clearpage

\section*{\Large{Appendix}}

\section{Comparison to implementing neural field autoencoder with LIIF~\cite{chen2021learning}}
\label{sec:cmp_liif_ae}

\begin{figure}
    \centering
    \includegraphics[width=\linewidth]{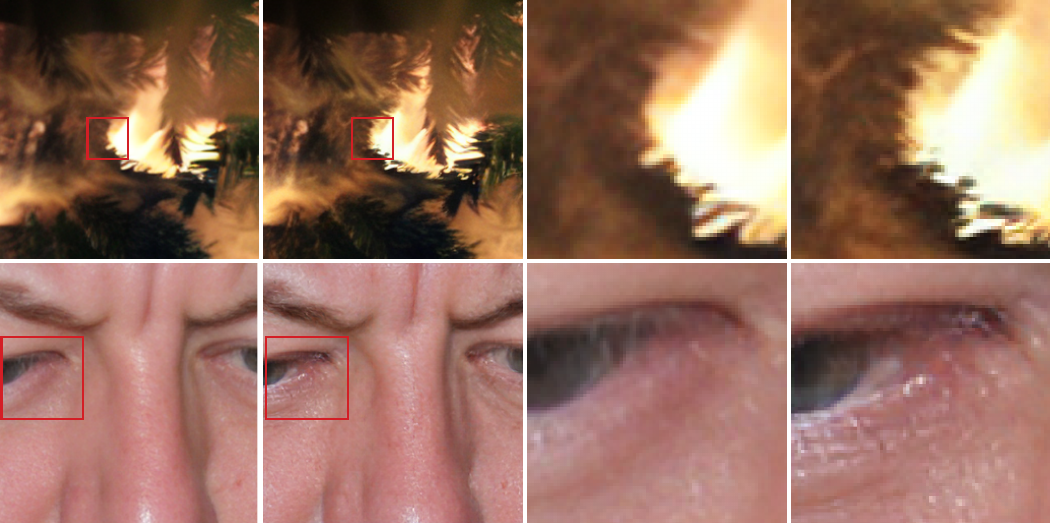}
    \caption{Comparison of using LIIF (left) and CLIF (right, ours) as the renderer in our framework (Mountains and FFHQ dataset). We find CLIF can produce more photorealistic details than LIIF with the convolutional formulation.}
    \label{fig:liif_vs_conv}
\end{figure}

LIIF~\cite{chen2021learning} is an image neural field defined on a feature map, which can also be used as the renderer in the framework. However, we observe that directly implementing our framework with LIIF does not produce photorealistic details, as shown in Figure~\ref{fig:liif_vs_conv}. LIIF was originally proposed for super-resolution with L1 loss. We find the generator with LIIF struggles to learn photorealistic details and the adversarial training quickly collapses with the discriminator as the winner in the adversarial game.

Our proposed CLIF renderer addresses this issue by decoding the patch as a whole and incorporating a larger context with convolution layers. Besides having higher capacity, we also find CLIF can be learned to be scale-consistent even with LPIPS and GAN loss and without point-independent decoding.

\section{Ablation on without latent space}
\label{sec:without_latent_space}

To evaluate the benefits of having a latent space for neural image fields, 
%
% or simply learning the any-scale super-resolution model is sufficient,
we compare our method, to a baseline ablation where we remove the encoder and latent space.
Instead, in this baseline, we first learn an any-scale super-resolution model for low-resolution images.
This super-resolution model has the same architecture as the composition of our decoder and renderer $R\circ D$.
% can be viewed as implementing a patch-based any-scale super-resolution model.
% which has the same architecture as our decoder and renderer,
The difference is that it acts on low-resolution images, rather than the latent representation.
We then train a low-resolution diffusion model.
At inference time, we first generate a low-resolution image by diffusion, then upsample it with $R\circ D$.
For this baseline, the low-resolution image has the same spatial dimensions 64$\times$64 as the latent representation in our main model, and all other training settings are kept identical.

\begin{figure}
    \centering
    \includegraphics[width=\linewidth]{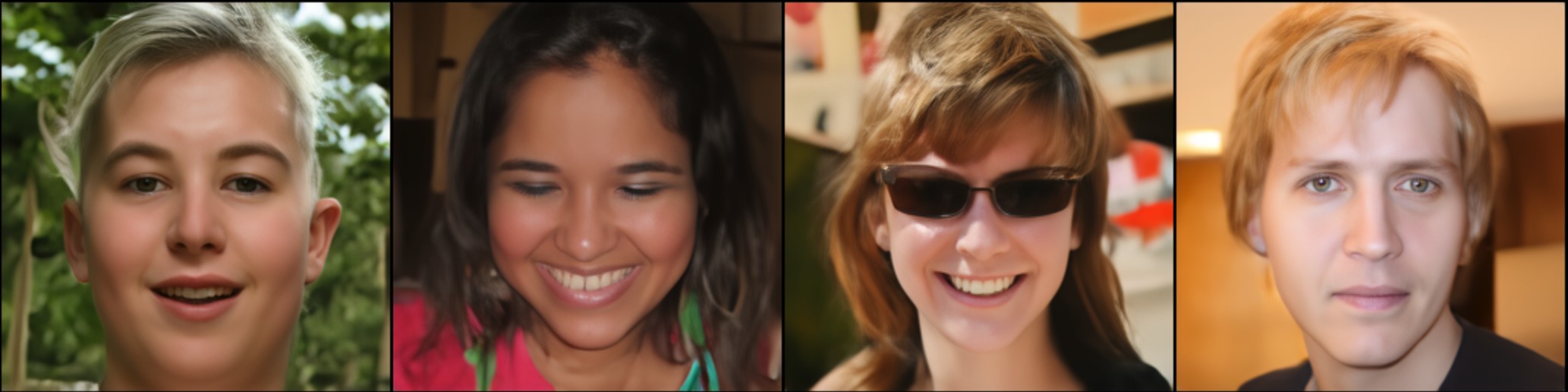}
    \caption{Samples of the method where the latent neural field space is replaced by low-resolution images with the same spatial dimension (FFHQ dataset). Even at 256 resolution, these images are overly smooth and lack details, which supports the effectiveness of having a latent space.}
    \label{fig:lrdmsr}
\end{figure}

Figure~\ref{fig:lrdmsr} shows this baseline generates overly smooth images.
This is already visible at 256$\times$256.
The quantitative comparison in Table~\ref{tab:lrdm-sr}, confirms our model outperforms the baseline on FID.
%
% The FID score is consistent with the visual results, which confirms that a low-resolution diffusion model with super-resolution is worse than our method.
%
We hypothesize this is because: (i) the latent representation contains richer information compared to a simple low-resolution RGB image with the same spatial dimension.
%, which will be synthesized by the powerful diffusion model. 
%
Intuitively, with the latent representation, our first training stage can encode information relevant to our final goal to synthesize a high-resolution image, which is impossible with a plain RGB low-resolution image.
This information is preserved by the diffusion model in the second training stage. (ii) The latent space (with VQ or KL regularization) is more robust than RGB space to the domain shift from real to generated sample.

% the diffusion model is aware that, so instead of only synthesizing a low-resolution image, it can directly synthesize the representation that corresponds to an image neural field and collaborate better with the decoder and renderer.

\begin{table}[]
    \centering
    \begin{tabular}{ccc}
        \toprule
        \textbf{Data} & \textbf{Method} & \textbf{FID-256@5K} \\
        \midrule
        \multirow{2}{*}{All HR} & LR-DM + SR & 36.01 \\
         & INFD & 9.26 \\
        \midrule
        \multirow{2}{*}{6K-Mix} & LR-DM + SR & 37.97 \\
         & INFD & 9.81 \\
        \bottomrule
    \end{tabular}
    \caption{Comparison to low-resolution diffusion model with any-resolution upsampler. The upsampling network has the same architecture as our decoder-renderer for comparison.}
    \label{tab:lrdm-sr}
\end{table}

\section{Convolutional samples of LDM~\cite{rombach2022high} for higher-resolution generation}
\label{sec:ldm_conv_samples}

\begin{figure}[]
    \centering
    \begin{tabular}{cc}
        \includegraphics[width=.45\linewidth]{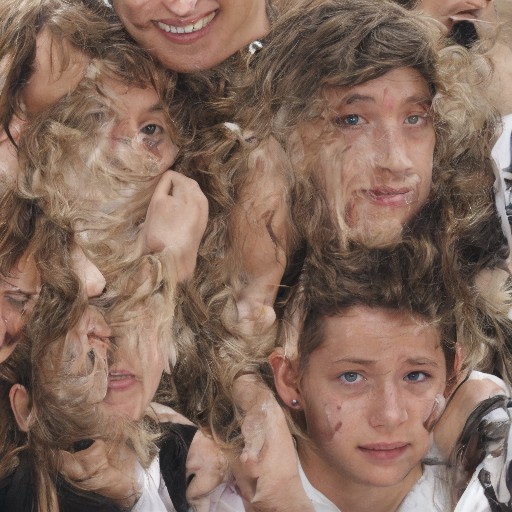} & \includegraphics[width=.45\linewidth]{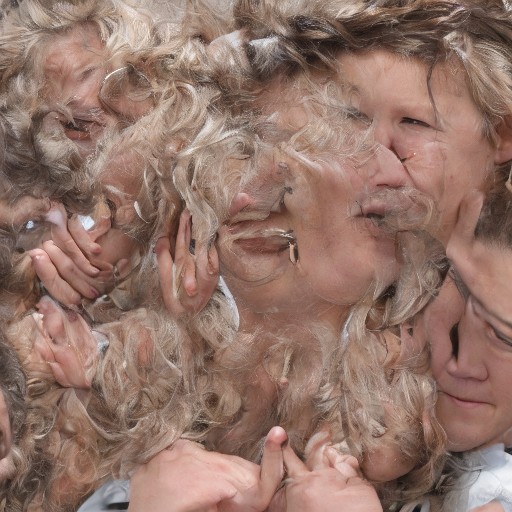}
    \end{tabular}
    \caption{Results using LDM~\cite{rombach2022high} trained for $256\times 256$ faces to generate higher-resolution faces at $512\times 512$ by making convolutional samples over a larger noise map.}
    \label{fig:ldm-fullconv}
\end{figure}

In LDM~\cite{rombach2022high}, the iterative denoising process is operating over a noise map with a UNet, where the UNet contains convolution and attention layers. Since both types of layers can be directly applied to higher-resolution input, a potential method to generate higher-resolution images is to process a noise map with higher resolution without changing the network. Despite this method increasing the computation cost linearly to the number of pixels, we observe that it fails to generate images with correct global structures, as shown in Figure~\ref{fig:ldm-fullconv} for the FFHQ unconditional generation task. We hypothesize this is because the model trained for $256\times 256$ images learns to generate faces at a specific scale, when making convolutional samples at $512\times 512$, it might try to generate multiple $256\times 256$ faces at different spatial locations and fails to preserve the global structure.

\begin{figure}
    \centering
    \includegraphics[width=\linewidth]{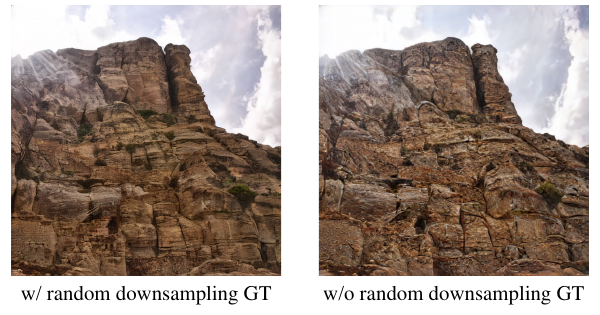}
    \caption{Comparison of training with/without random downsampling the ground-truth images (Mountains dataset, outputs in the autoencoder stage). Training without random downsampling produces samples with worse quality.}
    \label{fig:train_downsample}
\end{figure}

\section{Effect of scale-varied training on Mountains dataset}
\label{sec:scale_varied_mountains}

Scale-varied training, i.e., training with randomly downsampled ground-truth images, allows the same latent representation to get supervision from multiple scales with the fixed-resolution patch, which helps the performance even if all ground-truth images are at fixed and high resolution. Besides results on the FFHQ dataset shown in the main paper, we observe that scale-varied training is more important on the Mountains dataset which contains more complex images. As shown in Figure~\ref{fig:train_downsample}, without random downsampling of the ground-truth images during training, the model produces samples with worse quality.

\begin{figure}
    \centering
    \includegraphics[width=\linewidth]{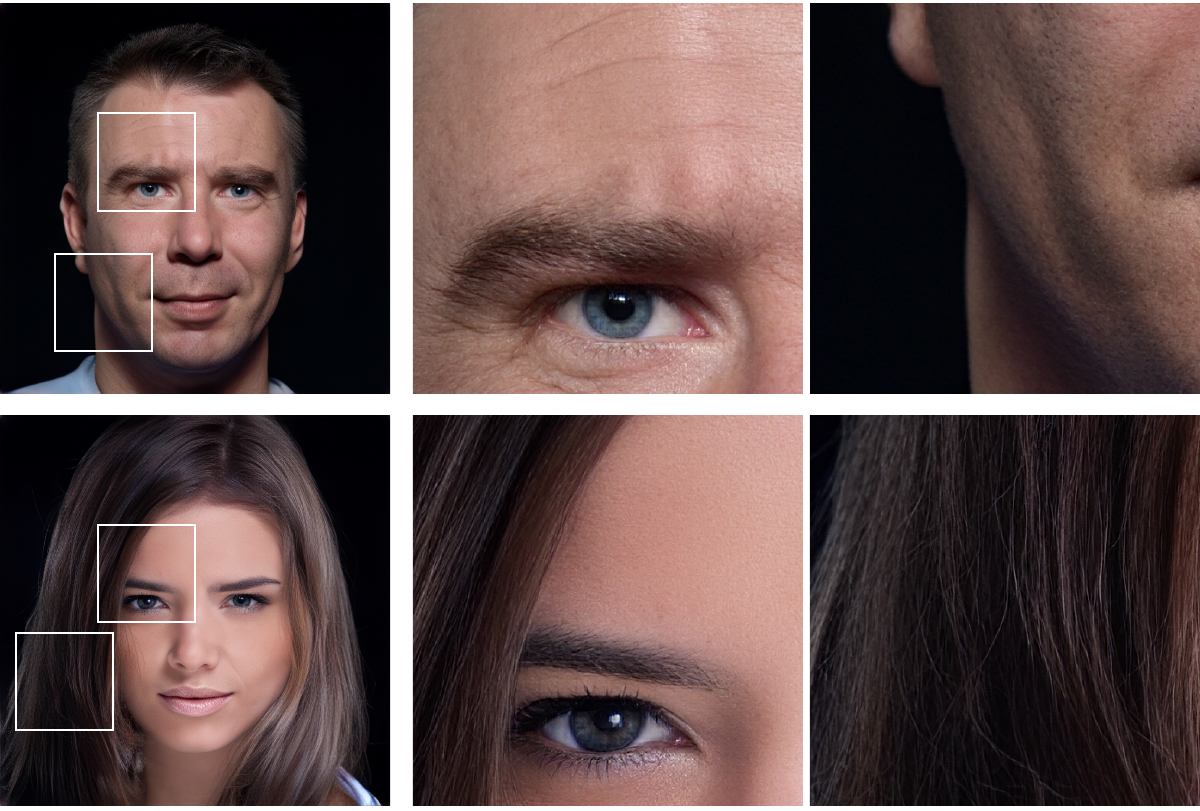}
    % \vspace{-.2in}
    \caption{{\bf 2K results on faces.} When training data is available, our method can be used to generate images at resolutions above 1K. Here we show 2K generated images trained on datasets of high-resolution portraits.}
    % \vspace{-1em}
    \label{fig:gsfaces}
\end{figure}

\section{Image generation beyond 1K}
\label{sec:gsfaces_results}

Our method uses patchwise supervision at arbitrary coordinates and can be trained on higher-resolution images without changing the architecture. 
We explore going beyond the 1024 resolution and train our model on a collected dataset of faces, which contains images at varied resolutions between 1K to 2K (about 84\% images have 2K resolution).
We show some qualitative results in Figure~\ref{fig:gsfaces}.
We observe that, with no change to the architecture, our method learns to generate highly detailed textures on skin and hairs for 2K faces, which suggests the potential for pushing our method further, for ultra-high resolution image generation.
% We also show 2K results on Mountains dataset in $\texttt{infd\_2k}/$.

\begin{table}[]
    \centering
    \small
    \begin{tabular}{cccc}
        \toprule
        \multirow{2}{*}{\textbf{Model}} & \multicolumn{3}{c}{\textbf{pFID@50K}} \\
        \cmidrule(lr){2-4}
         & 256/1K & 512/1K & 1K/1K \\
        \midrule
        AnyRes-GAN~\cite{chai2022any} & 6.17 & 4.02 & 3.25 \\
        INFD & 7.53 & 6.84 & 5.13 \\
        \bottomrule
    \end{tabular}
    \caption{FID comparison between any-resolution GAN and diffusion model on Mountains dataset. While GANs are state-of-the-art at single class FID, the diffusion-based method achieves competitive FID and does not have the GAN artifacts shown in Figure~\ref{fig:anyresgan}, and shows better visual quality on actual images. We refer to Sec.~\ref{sec:cmp_anyresgan} for random samples and detailed discussions.}
    \vspace{-1.5em}
    \label{tab:cmp_anyres}
\end{table}

\section{Discussion on dataset scale-consistency}
\label{sec:dataset_scale_consistency}

Our model assumes the images in the dataset to be scale-consistent, i.e., downsampled high-resolution images follow the same distribution as low-resolution images. Datasets that severely violate this assumption would degrade the performance of our model. This is because if dataset scale consistency is violated, the latent code encoded from low-resolution images might follow a different distribution than the latent code encoded from downsampled high-resolution images. The high-resolution supervision is only applied to the latter type of latent code during training, therefore, the former type of latent code might not have a guaranteed quality when rendered at high resolution. For text-to-image synthesis, our current model is finetuned with clean high-resolution data from the Stable Diffusion model, while the Stable Diffusion model could have seen many noisy low-resolution images in its pre-training, appending text prompts like ``high definition'' or ``4k'' reduces such distribution shift and thus improves the quality. Training from scratch with scale-consistent data will not have a distribution shift in the latent space and might thus avoid this issue.

\section{Additional Generated Samples}
\label{sec:additional_samples}

\begin{figure*}
    \centering
    \includegraphics[width=\linewidth]{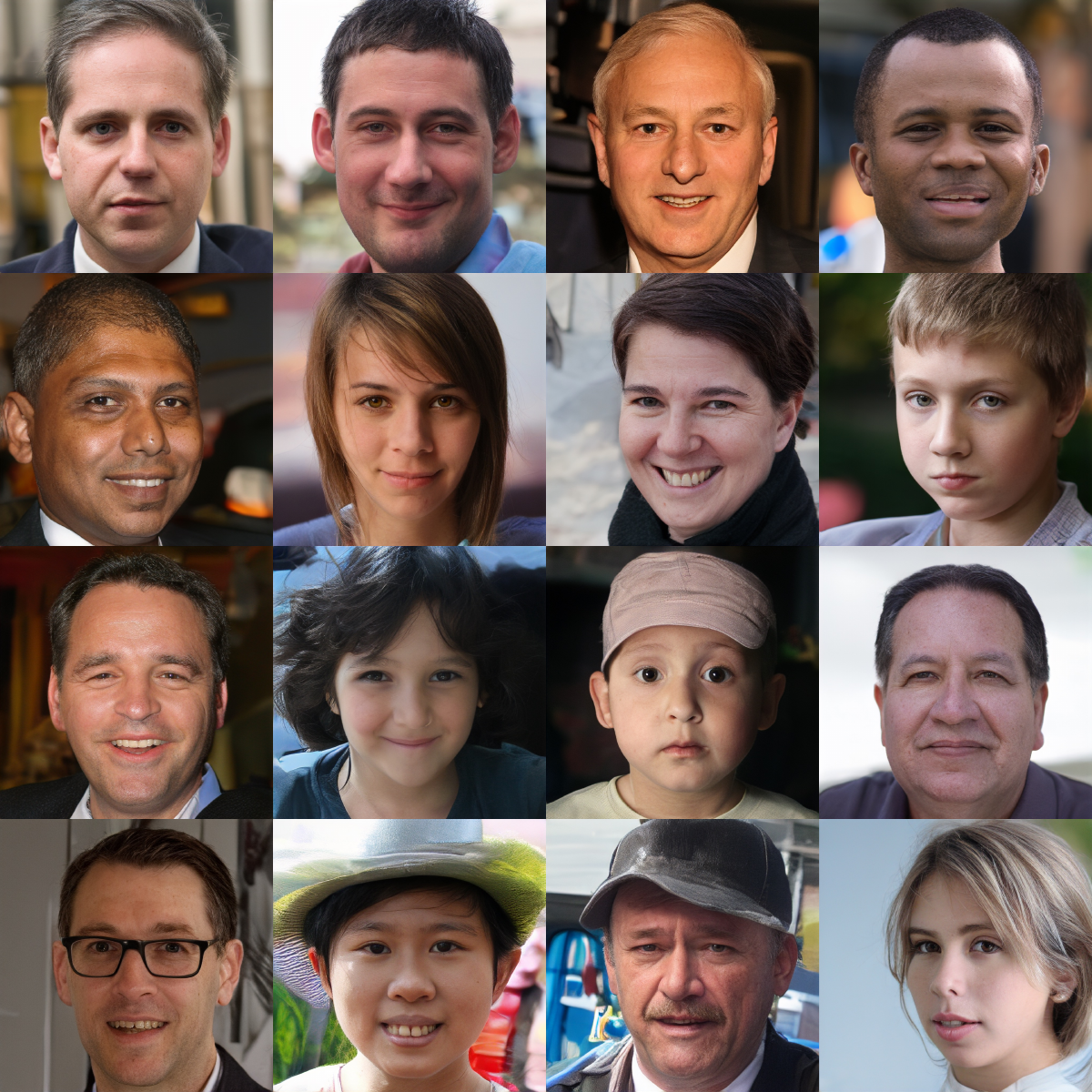}
    \caption{Additional generated samples of our method on FFHQ-1024.}
    \label{fig:add_ffhq}
\end{figure*}

\begin{figure*}
    \centering
    \begin{tabular}{ccc}
        \includegraphics[width=.45\linewidth]{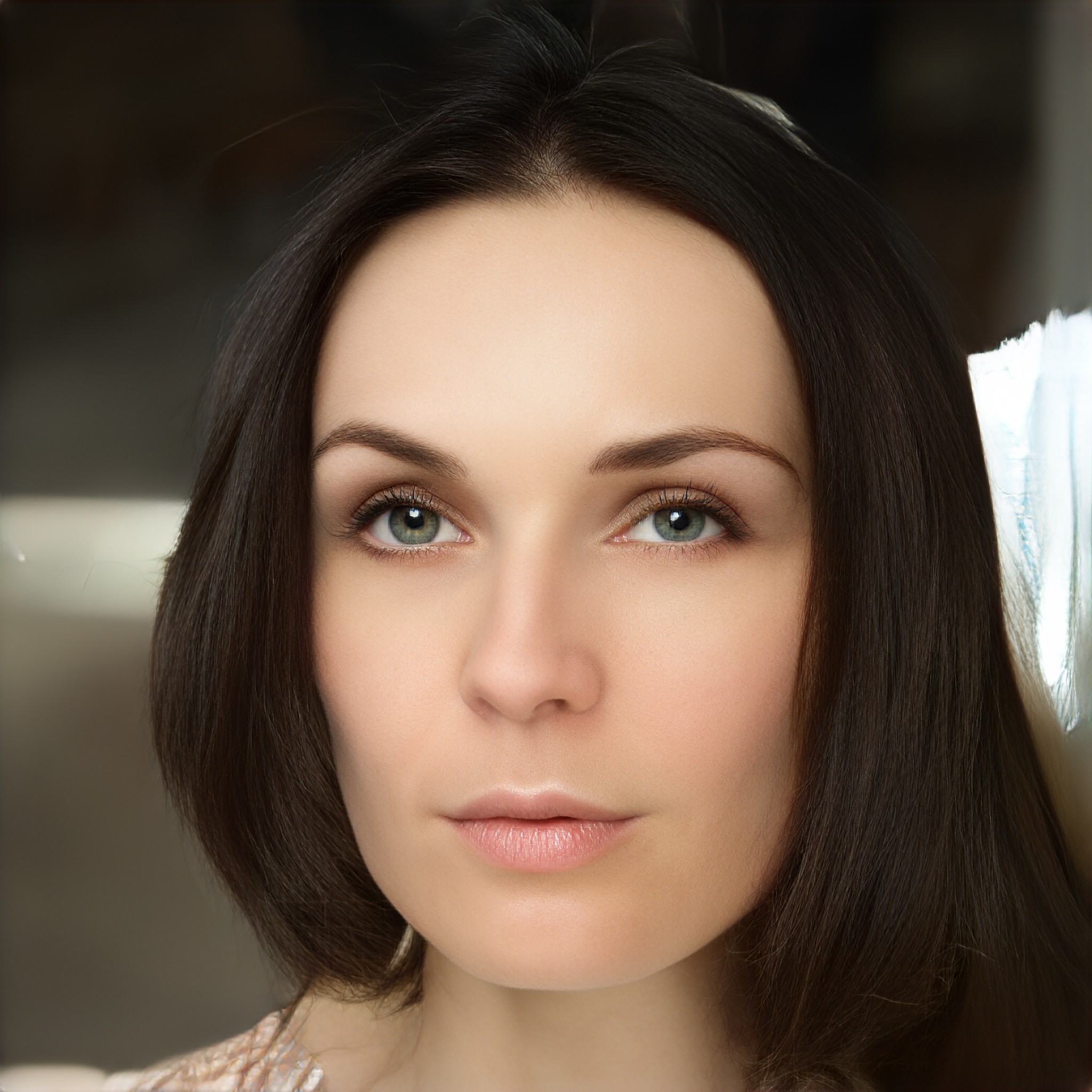} & \includegraphics[width=.45\linewidth]{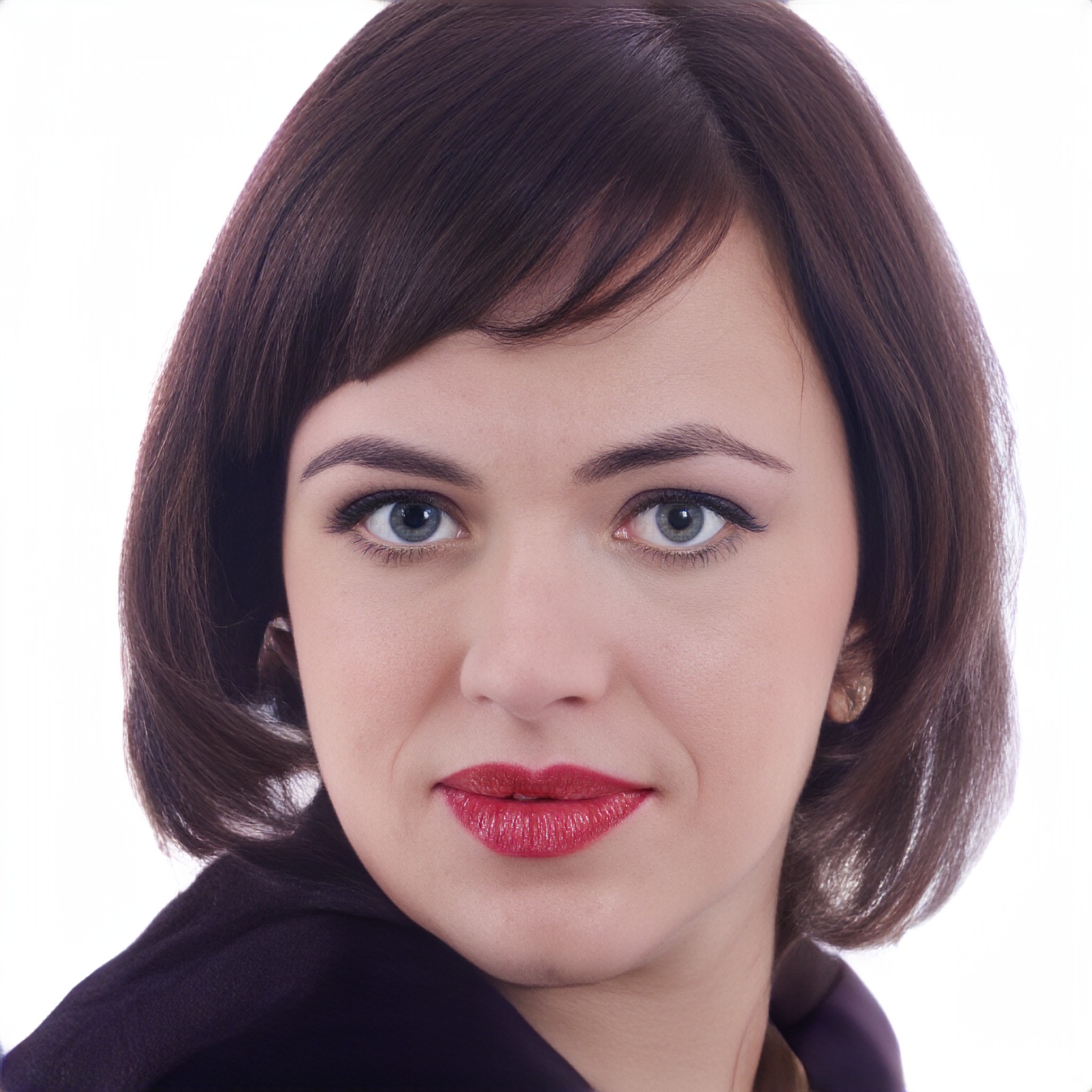} \\
        \includegraphics[width=.45\linewidth]{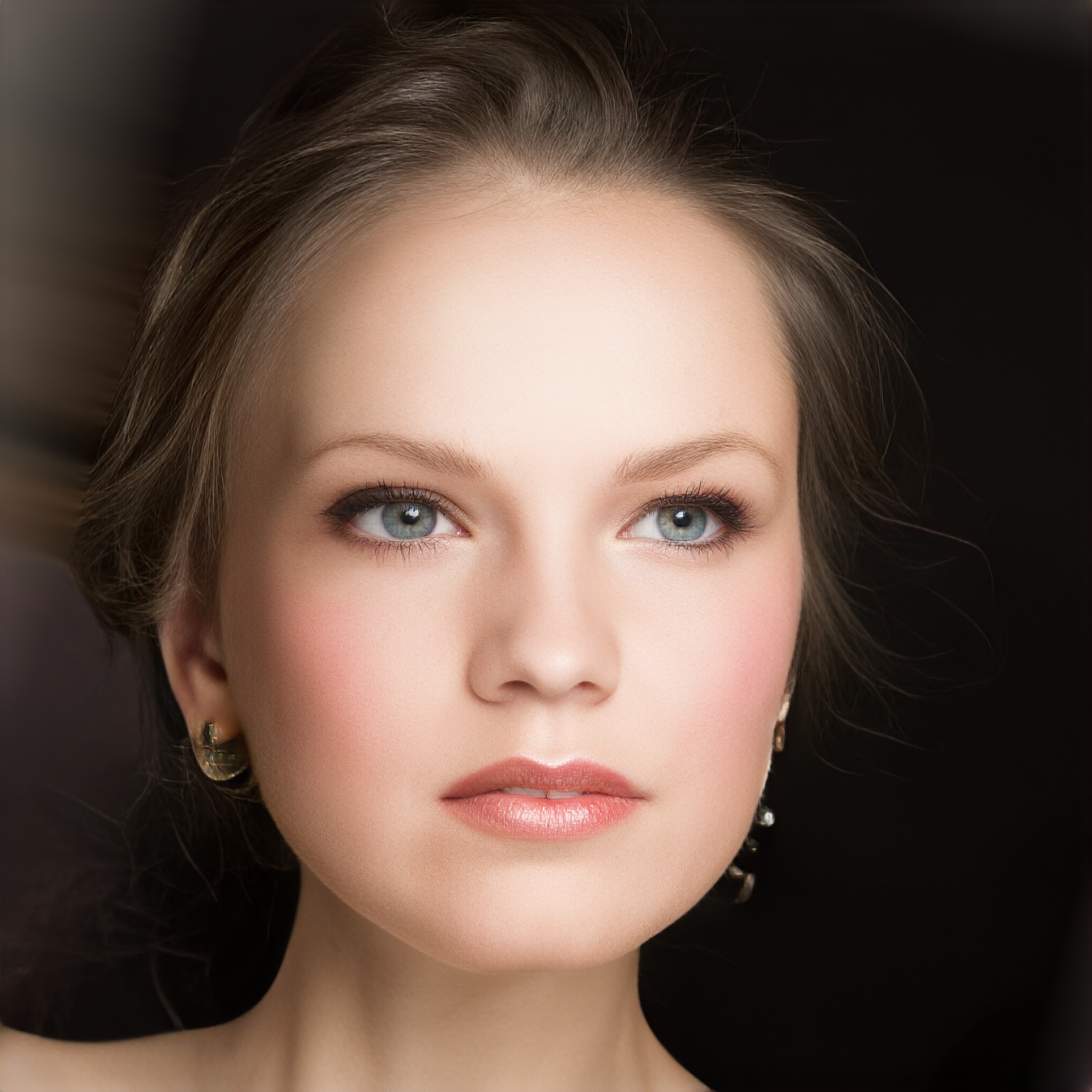} & \includegraphics[width=.45\linewidth]{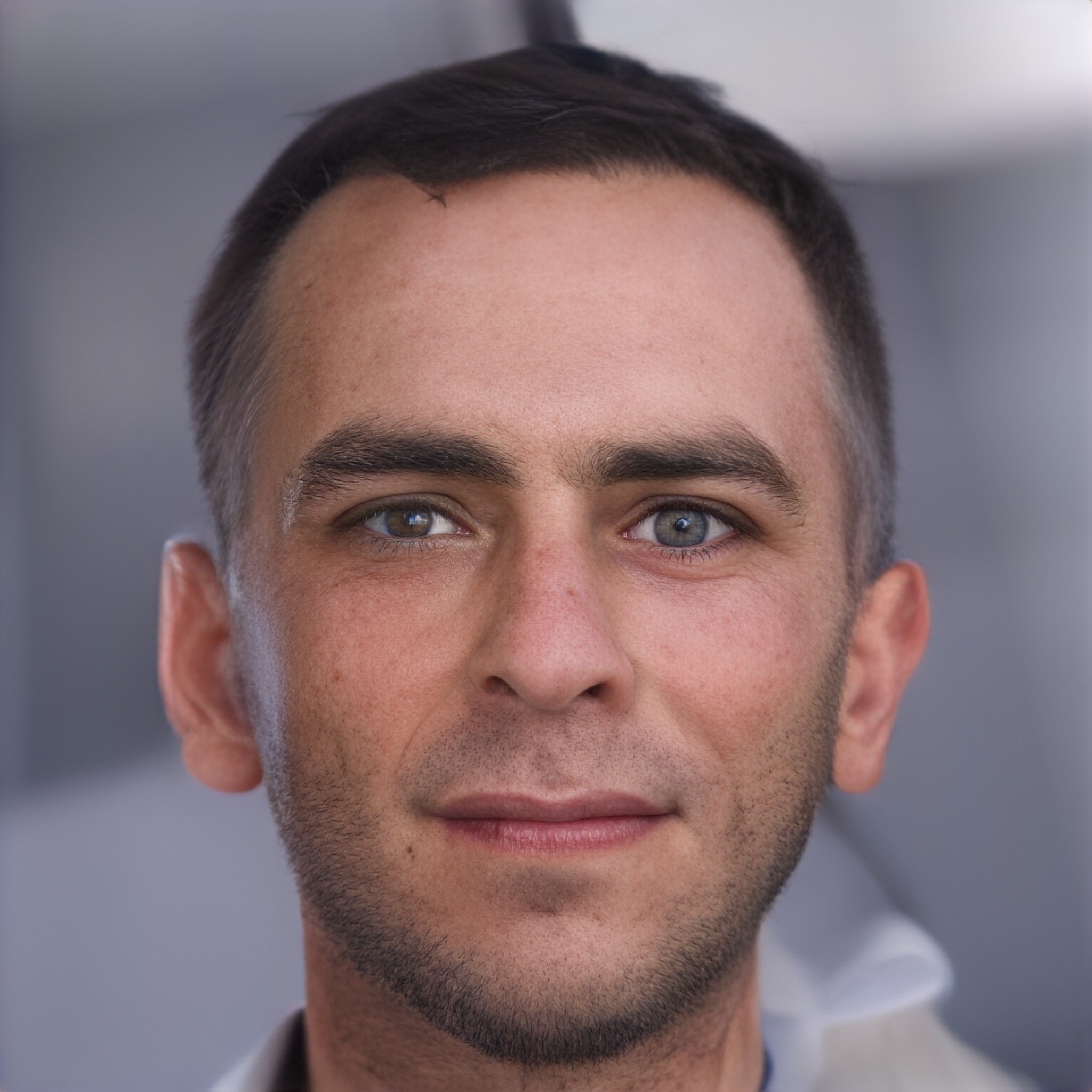}
    \end{tabular}
    \caption{Additional generated samples of our method on 2K portrait dataset.}
    \label{fig:add_gsfaces}
\end{figure*}

\begin{figure*}
    \centering
    \includegraphics[width=\linewidth]{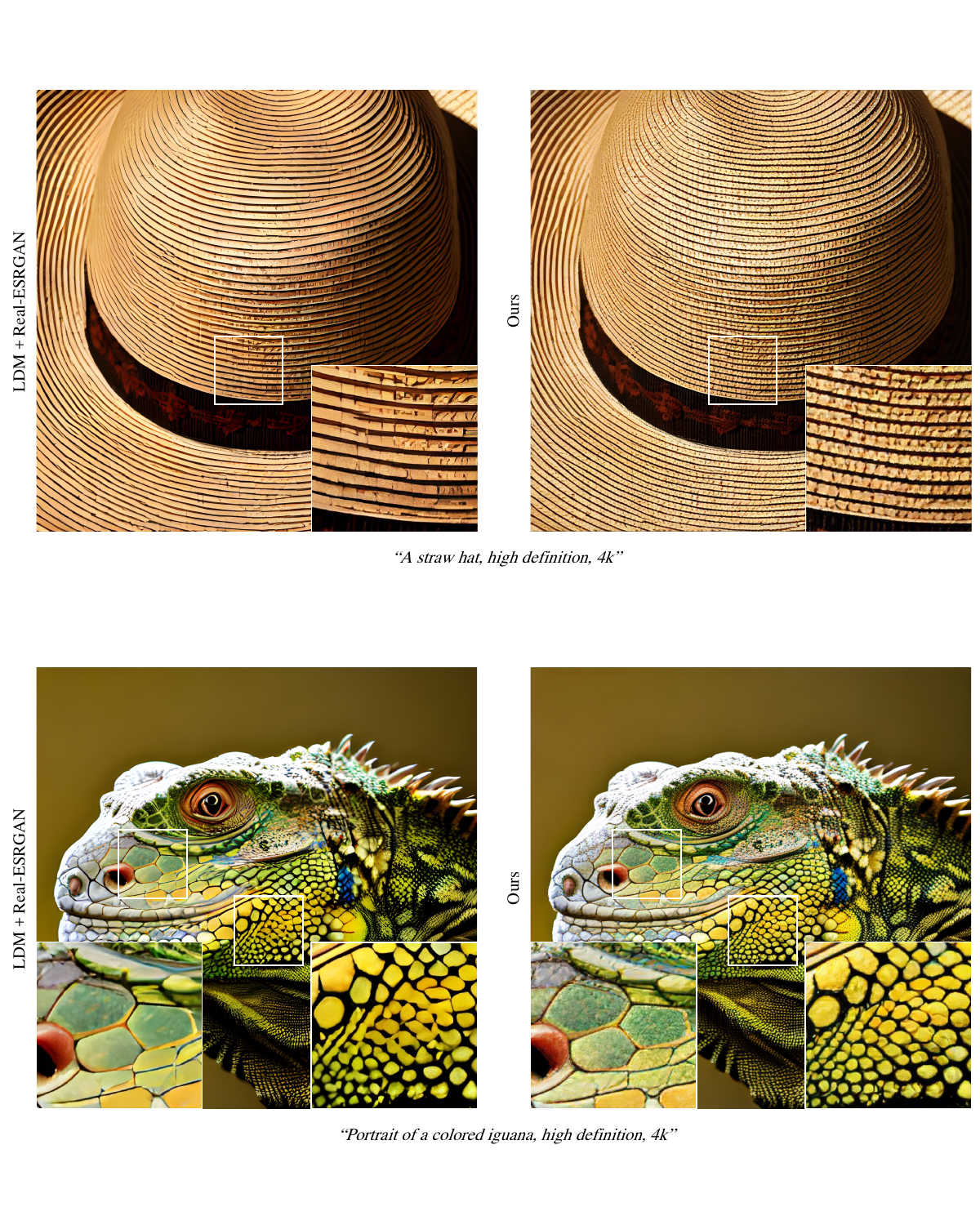}
    \caption{Additional generated samples of our method on text-to-image generation (resolution at 2K).}
    \label{fig:add_txt2img2}
\end{figure*}

\begin{figure*}
    \centering
    \includegraphics[width=\linewidth]{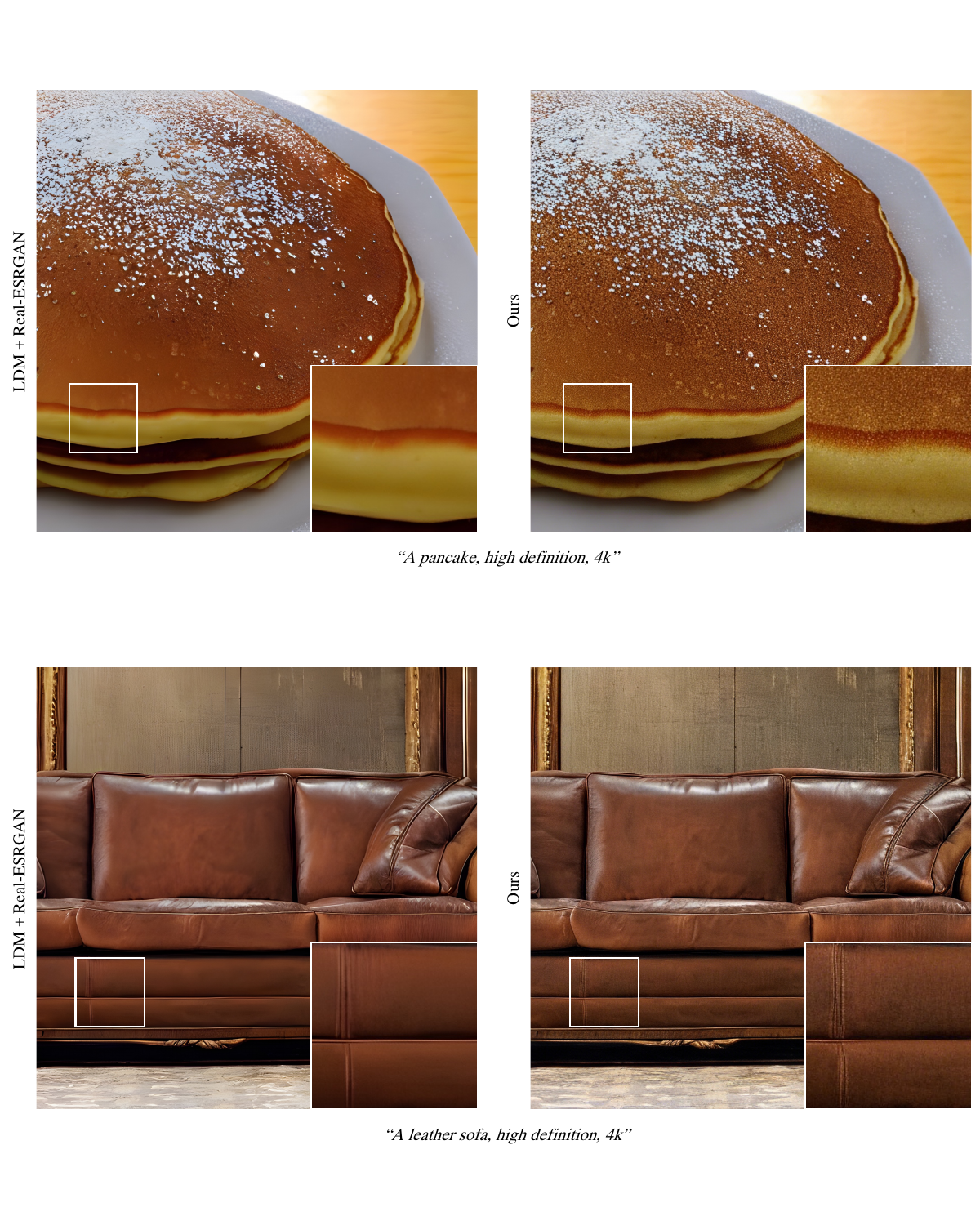}
    \caption{Additional generated samples of our method on text-to-image generation (resolution at 2K).}
    \label{fig:add_txt2img3}
\end{figure*}

\begin{figure*}
    \centering
    \includegraphics[width=\linewidth]{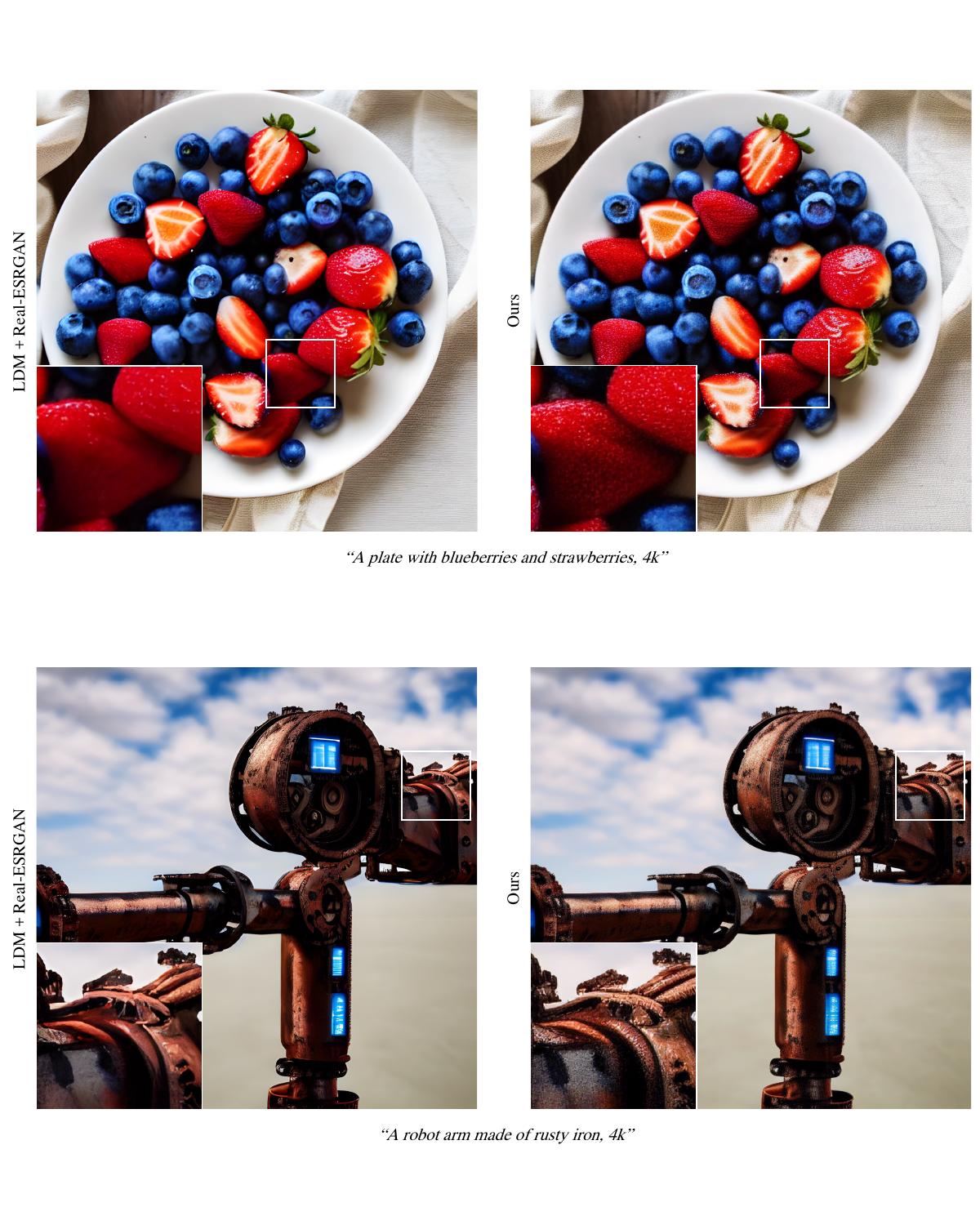}
    \caption{Additional generated samples of our method on text-to-image generation (resolution at 2K).}
    \label{fig:add_txt2img1}
\end{figure*}

We show various additional generated samples in Figures~\ref{fig:add_ffhq},\ref{fig:add_gsfaces},\ref{fig:add_txt2img2},\ref{fig:add_txt2img3},\ref{fig:add_txt2img1} for FFHQ-1024, 2K portrait dataset, and text-to-image generation at 2K resolution, where the experimental settings are the same as the main paper. In text-to-image generation, we observe that while the output resolution is at 2K, appending the prompt ``high definition'' or ``4k'' after the text description is helpful to generate high-quality high-resolution images.

\begin{figure*}
    \centering
    \includegraphics[width=.8\linewidth]{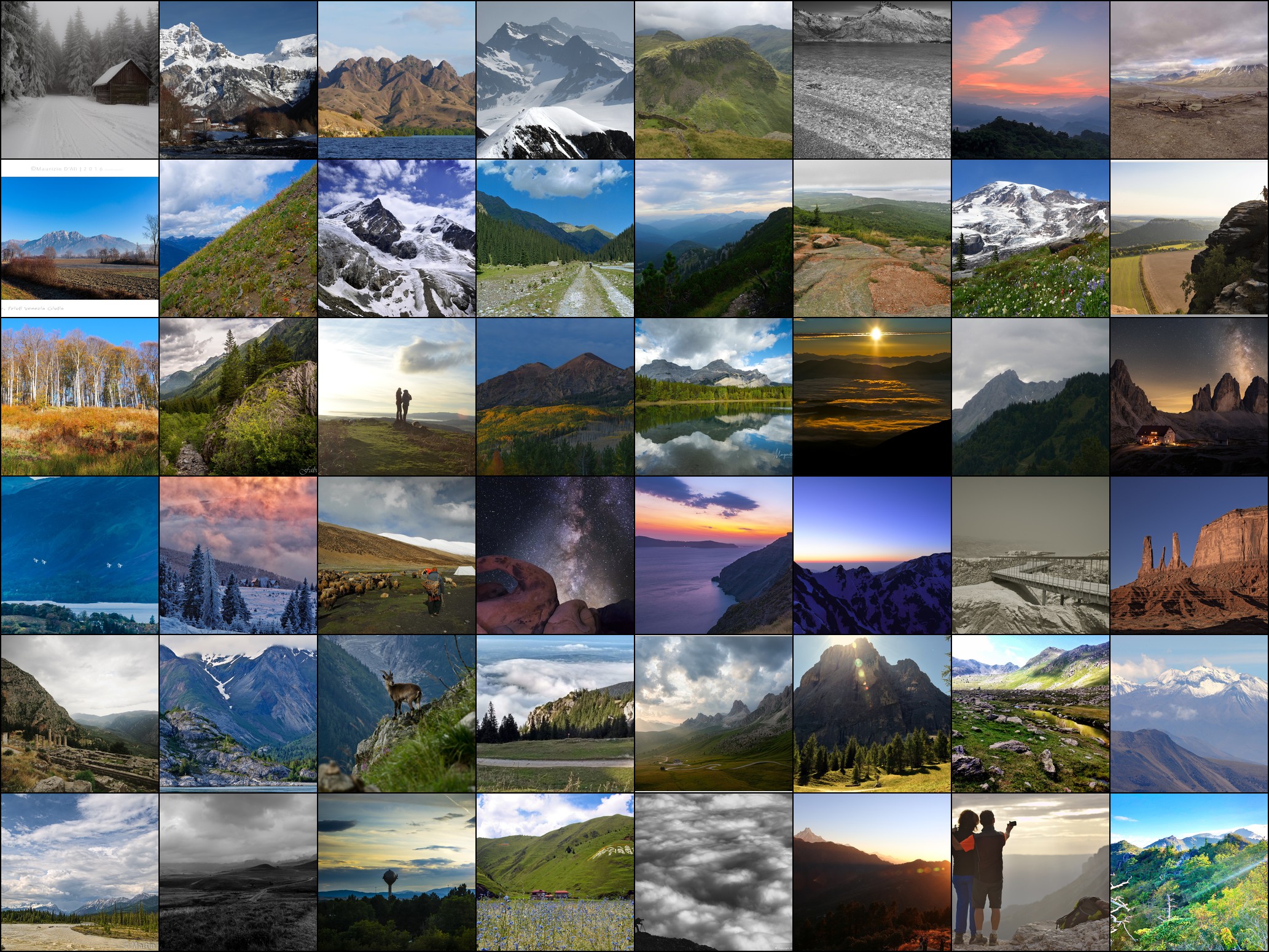}
    \caption{Sample diversity of ground-truth images on Mountains dataset (shown in $256\times 256$).}
    \label{fig:sample_diversity_gt}
\end{figure*}

\begin{figure*}
    \centering
    \includegraphics[width=.8\linewidth]{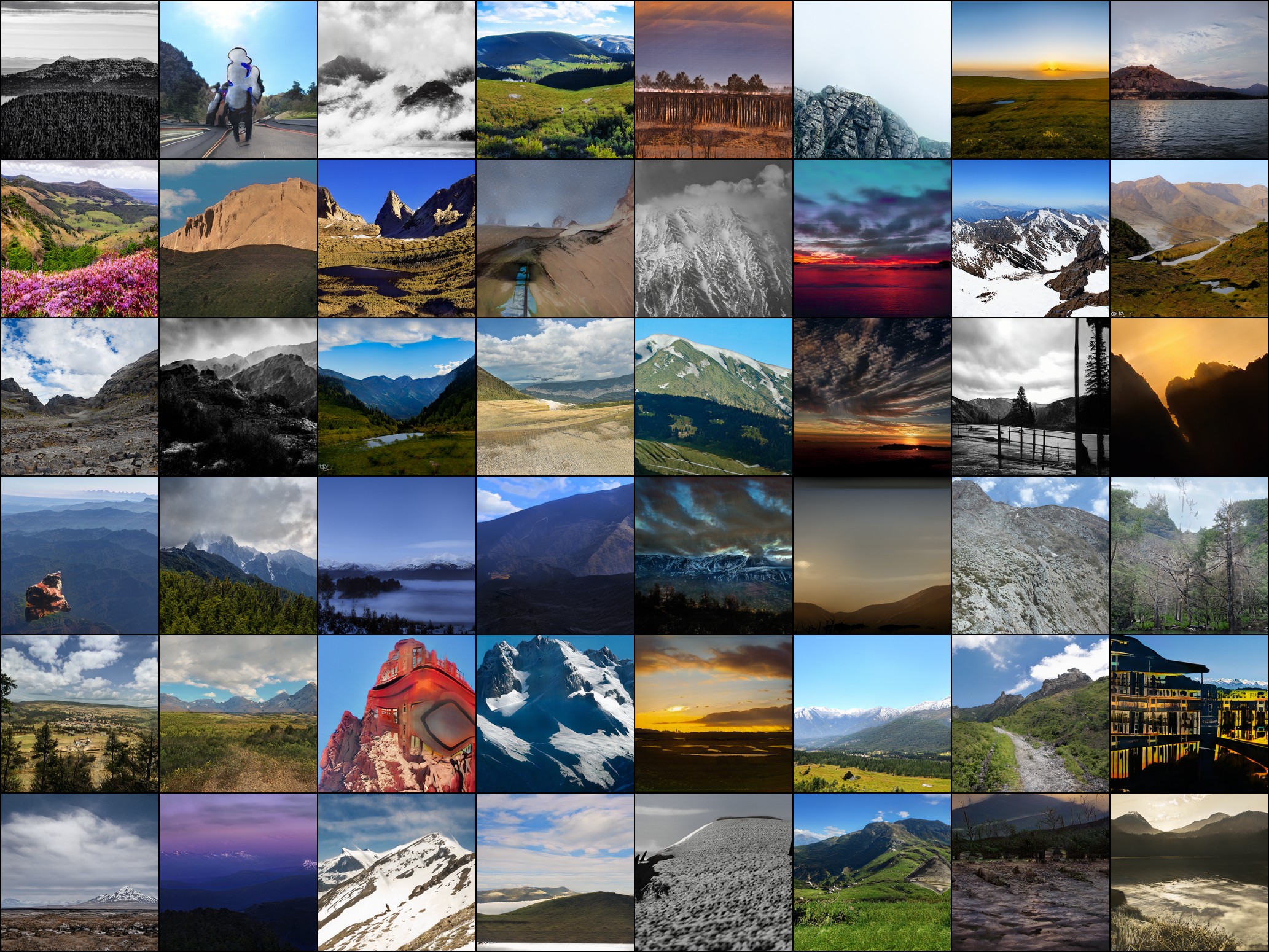}
    \caption{Sample diversity of AnyResGAN~\cite{chai2022any} on Mountains dataset (shown in $256\times 256$).}
    \label{fig:sample_diversity_gan}
\end{figure*}

\begin{figure*}
    \centering
    \includegraphics[width=.8\linewidth]{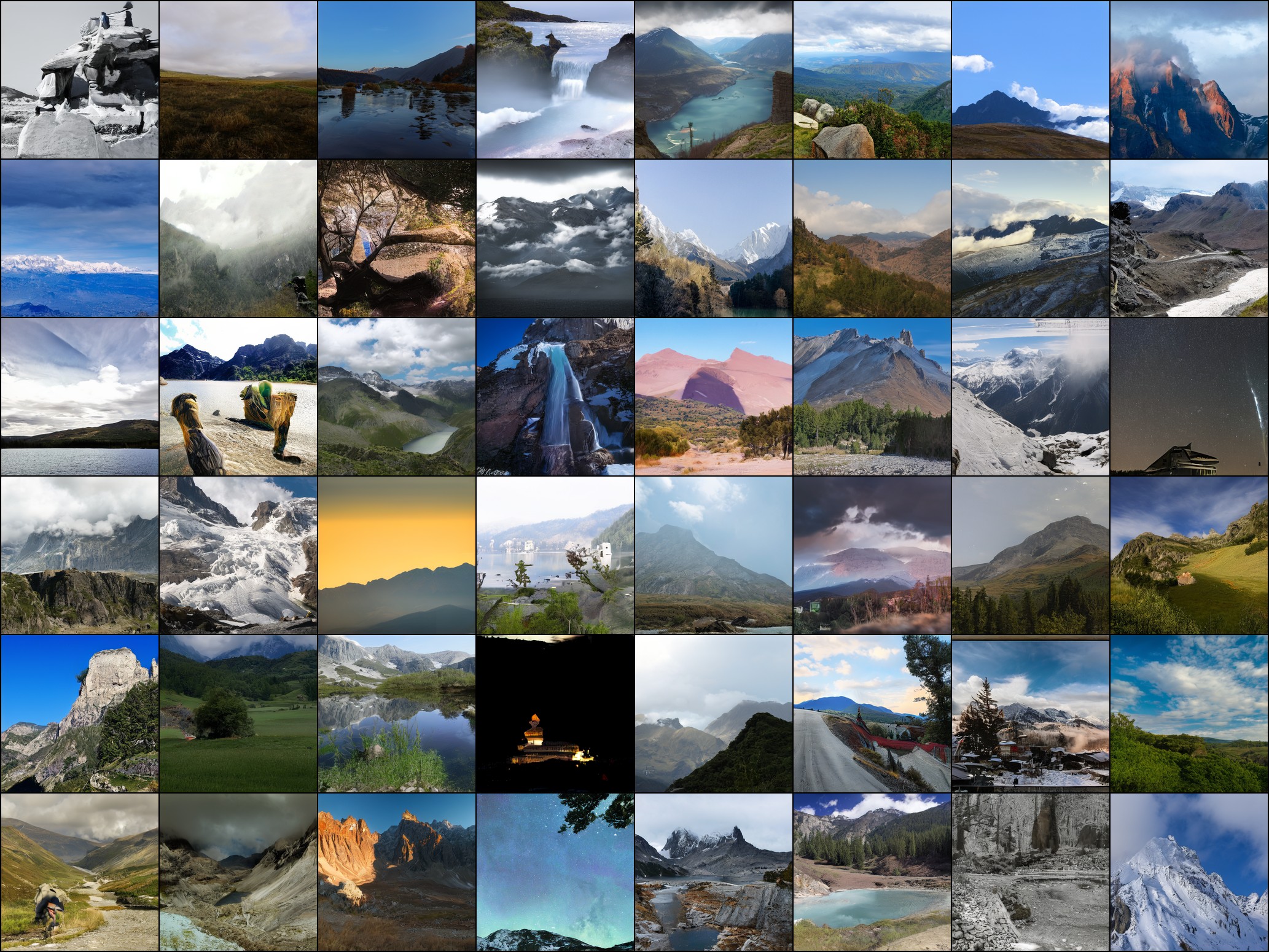}
    \caption{Sample diversity of image neural field diffusion model on Mountains dataset (shown in $256\times 256$).}
    \label{fig:sample_diversity_dm}
\end{figure*}

\section{Implementation Details}

\subsection{Model architecture} Our encoder and decoder follow the architecture used in LDM~\cite{rombach2022high}. They are modified from a UNet's encoder and decoder by removing the connections that skip the bottleneck latent representation. The encoder and decoder are symmetric, each has 3 levels. Downsampling/upsampling happens after each level. The base channel is 128, the channel multiplication factors are 1,2,4 for different levels in the encoder. There are 2 ResNet blocks within each level. The feature map of latent representation has a downsampling rate of 4 compared to the input and has 3 channels. The CLIF renderer is a convolutional neural network with one convolution layer, and two ResNet blocks, followed by another convolution layer, convolution kernel sizes are all 3.

Our diffusion model in latent space follows the implementation in ADM~\cite{dhariwal2021diffusion} (also used in LDM~\cite{rombach2022high}). The encoder and decoder have base channels 224 and channel multiplication factors are 1,2,3,4 at different encoder levels, with 2 ResNet blocks at every level. At the downsampling rates of 2,4,8, multi-head self-attention with 32 channels per head is applied on the feature map.

\subsection{Training setting}

For the first stage, the encoder, decoder, and renderer are end-to-end trained jointly. We use Adam~\cite{kingma2014adam} with $\beta_1=0.5,\beta_2=0.9$ and optimize for 1M iterations. The learning rate is $3.6\cdot 10^{-5}$ for a batch size of 8. The discriminator for GAN loss is adversarially trained with the same optimizer specifications. On Mountains and text-to-image tasks, we keep the ground-truth images at their original resolution with a probability of 0.5 in the last 400K iterations.

For the second stage, the latent space diffusion model is trained with AdamW~\cite{loshchilov2017decoupled} for 600K iterations on FFHQ, for 1.7M iterations on Mountains, with $\beta_1=0.9,\beta_2=0.999$ and weight decay of $0.01$. The learning rate is $9.6\cdot 10^{-5}$ for a batch size of 48. For either the first stage or the second stage, it takes about 4 days every 1M iterations to train our model on 4 NVIDIA A100 GPUs.

% WARNING: do not forget to delete the Supplementary Material pages from your submission 
% \input{sec/X_suppl}

\end{document}